\documentclass[]{fairmeta} %

\usepackage{graphicx} %
\usepackage{comment}
\usepackage{palatino}
\usepackage{pifont}
\usepackage{graphicx}
\usepackage{amsmath}
\usepackage{amsfonts}
\usepackage{enumitem}
\usepackage{booktabs}
\usepackage[dvipsnames]{xcolor}
\usepackage{colortbl}
\usepackage{xspace}
\usepackage[font=small]{caption}
\newcommand{\xmark}{\ding{55}}%
\newcommand{\cmark}{\ding{51}}%
\usepackage{titlesec}
\usepackage{longtable}
\usepackage{booktabs}
\newcommand{\cReLU}{\textcolor{blue}{\relu}}
\newcommand{\cSiLU}{\textcolor{orange}{\silu}}

\newcommand{\cTANH}{\textcolor{brown}{\tanh}}

\title{Stochastic activations}

\author[1]{Maria Lomeli}
\author[1]{Matthijs Douze}
\author[1]{Gergely Szilvasy}
\author[1,2,3]{Loic Cabannes}
\author[1]{Jade Copet}
\author[1]{Sainbayar Sukhbaatar}
\author[1]{Jason Weston}
\author[1]{Gabriel Synnaeve}
\author[1]{Pierre-Emmanuel Mazar\'e}
\author[1]{Herv\'e J\'egou}
\affiliation[1]{Meta FAIR}
\affiliation[2]{Ecole Normale Supérieure Paris Saclay}
\affiliation[3]{Paris Cité University}
\abstract{We introduce stochastic activations. This novel strategy randomly selects between several non-linear functions in the feed-forward layer of a large language model.
In particular, we choose between \silu or \relu depending on a Bernoulli draw. This strategy circumvents the optimization problem associated with \relu, namely, the constant shape for negative inputs that prevents the gradient flow. 
We leverage this strategy in two ways:

(1) We use stochastic activations during pre-training and fine-tune the model with \relu, which is used at inference time to provide sparse latent vectors. This reduces the inference FLOPs and translates into a significant speedup on CPU and GPU. This leads to better results than training from scratch with the \relu activation function. 

We evaluate stochastic activations for sequence generation. This strategy performs reasonably well: it has higher diversity and has only slightly inferior performance to the best deterministic non-linearity, \silu, combined with temperature sampling. This provides an alternative way to increase the diversity of  generated text.
}

\usepackage{xcolor}
\definecolor{darkgreen}{rgb}{0.0, 0.8, 0.13} %

\makeatletter
\DeclareRobustCommand\onedot{\futurelet\@let@token\@onedot}
\def\@onedot{\ifx\@let@token.\else.\null\fi\xspace}
 
\def\ie{\emph{i.e}\onedot}

\makeatother

\def\relu{\texttt{RELU}\xspace}

\def\silu{\texttt{SILU}\xspace}
\def\selu{\texttt{SELU}\xspace}
\def\gelu{\texttt{GELU}\xspace}
\def\tanh{\texttt{TANH}\xspace}
\def\swish{\texttt{Swish}\xspace}
\def\swiglu{\texttt{SWIGLU}\xspace}
\def\ash{\texttt{ASH}\xspace}
\def\rmsp{\texttt{R-S+}\xspace}

\def\swift{\texttt{Swi+FT}\xspace}
\def\stocha{\texttt{StochA\xspace}\xspace}
\def\tanhft{\texttt{Tanh+FT}\xspace}

\def\amaiasmall{\texttt{LM1.5B}\xspace} %
\def\amaiamed{\texttt{LM3B}\xspace}
\def \stosilu{\texttt{[S|R]-R+}\xspace}
\def\stodoublesilu{\texttt{[S|R]-S+}\xspace}
\def\stodoubletanh{\texttt{[T|R]-T+}\xspace}
\newcommand{\drawat}[3]{\makebox[0pt][l]{\raisebox{#2}{\hspace*{#1}#3}}}
\newcommand{\addstar}[1]{\phantom{$^\star$}#1$^\star$}

\date{December 23\textsuperscript{rd} 2025}

\begin{document}
\maketitle

\bigskip 

\begin{center}
\includegraphics[width=0.47\linewidth]{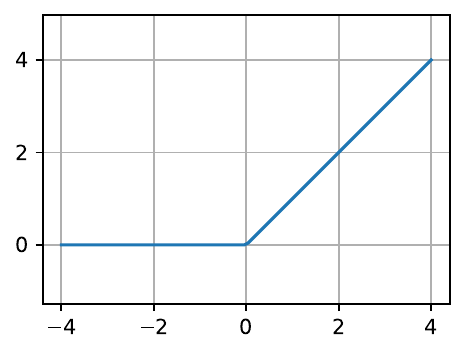}%
\drawat{-7cm}{5.2cm}{\colorbox{white}{\framebox{\relu: $1-p$}}}
\hfill
\includegraphics[width=0.47\linewidth]{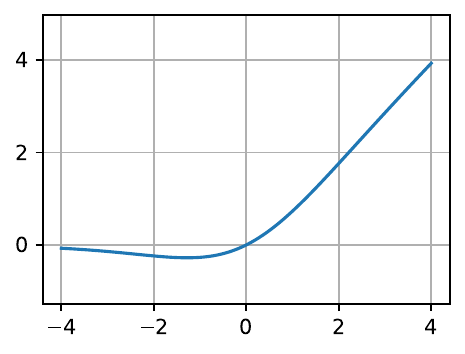}%
\drawat{-7cm}{5.2cm}{\colorbox{white}{\framebox{\silu: $p$}}}
\captionof{figure}{Stochastic activation randomly selects one of two activations when $x<0$: \newline (1) \relu selected with probability $1-p$; otherwise (2) another activation, in particular \silu.
\label{fig:bernoulli}
}
\end{center}
\bigskip

\section{Introduction}
\label{sec:introduction}

Large language models (LLMs)~\citep{devlin2019bertpretrainingdeepbidirectional,chowdhery2022palmscalinglanguagemodeling,brown2020language, Vaswani2017AttentionIA} have revolutionized natural language processing, enabling unprecedented capabilities in text generation, comprehension, and reasoning. Their success stems from scaling model parameters and leveraging vast amounts of data, but this comes with a significant computational complexity. As the demand for more efficient and powerful models grows, researchers are increasingly focused on optimizing their training processes to balance performance with resource constraints. 

The majority of the LLM parameters are in the Feed-Forward Network (FFN) layers, where they primarily serve to store and recall information from the training data.
FFNs are two linear layers separated by an \emph{activation function}, and sometimes an additional linear layer that serves as a gating operation.  
The activation is a non-linear function from $\mathbb{R}$ to $\mathbb{R}$. %
In this context, the choice of activation function plays a crucial role for both the model's expressivity and efficiency. 
The simplest choice of activation is \relu (Rectified Linear Unit), that allows positive values to pass through and forces negative inputs to zero. 
\relu is \emph{sparsity-inducing}, since, on average, half of its outputs are zero (in practice significantly more). 
Within a two-layer Multilayer Perceptron, this means that inference on the second layer is a matrix-sparse vector multiplication, hence, it can be implemented with fewer FLOPs than a matrix-dense vector multiplication. 
Note that effectively improving the runtime with this sparsity pattern remains challenging. 

In practice, the Sigmoid Linear Unit (\silu) activation, combined with a gated design, has consistently outperformed \relu in terms of model accuracy \citep{Shazeer2020GLUVI}. 
Unfortunately, \silu does not induce sparsity. 
One plausible explanation for \relu's underperformance is that its gradient for negative input values is zero, which hinders optimization by preventing weight updates in a significant portion of the network. 
Solutions like Leaky \relu \citep{maas2013rectifier} circumvent this problem by ensuring non-zero gradient almost everywhere, but they are inferior to \silu and involve abandoning sparsity. 
In contrast, if the so-called ``dying \relu problem'' optimization challenge could be effectively addressed, {\relu}’s theoretical advantages -- such as sparsity and computational efficiency -- may translate into performance comparable to \silu for a lower number of FLOPs. This disparity presents a challenge: how to harness the efficiency benefits of sparse activations, such as \relu, without sacrificing the empirical advantages of \silu. This motivates our exploration of alternative training strategies that mitigate {\relu}’s limitations while preserving its benefits.

In this work, we consider two ways to approach this problem. The first approach is activation fine-tuning, denoted by \swift: we pre-train the model with an activation that facilitates efficient large language model optimization, then, we change the activation to \relu and adapt the model by fine-tuning it further. 
Our second approach, referred to as \stocha (stochastic activations), is a novel technique that randomly selects between multiple activations, either at train or test time. 
Both approaches allow models to benefit from the superior optimization properties of \silu. These hybrid strategies combine the best of both worlds -- maintaining high model performance while unlocking the computational efficiency of sparse activations. 

In summary, this paper makes the following contributions: 
\begin{itemize}
    \item We introduce two strategies that employ different activation functions at training and inference time, namely \swift and \stocha. Both are complementary and make it possible to use activations at inference time that differ from those employed during pre-training. 
    \item We produce \relu-based models that are much better than those obtained with regular training, i.e., our methods significantly outperform training with \relu only. 
    \item We show that stochastic activations, when used at inference time, provide an alternative way to generate diverse sequences compared to traditional temperature sampling or other variants. %
\end{itemize}

\section{Related work}
\label{sec:related}

\paragraph{Standard activation functions} Activations are at the core of deep learning, they are fundamental to enable deep learning models to go beyond linear function with limited expressivity. 
While early neural network architectures were inspired by logistic regression, such as sigmoidal and tanh activations, many activation functions have been evaluated for the FFN layers of transformers. ~\citet{Vaswani2017AttentionIA} used 
\relu~\citep{glorot2011deep} initially. However, using the \relu activation function leads to some neurons getting stuck in the negative region. As a consequence, models stop learning entirely, since the gradient is zero for negative inputs, and their weights do not get updated. In contrast,~\citet{Touvron2023LLaMAOA} used \silu as the activation function for the FFN layers of the transformer for the first Llama models.~\citet{Shazeer2020GLUVI} discusses the benefits of \swiglu, which consists of \silu with gating.
There exist many other activation functions such as
the Gaussian Error Linear Unit (\gelu)~\citep{Hendrycks2016GaussianEL}, Scaled Exponential Linear Unit (\selu)~\citep{Klambauer2017SelfNormalizingNN}, \swish~\citep{Ramachandran2018SearchingFA} and gated \silu, among others.

In particular, the Leaky \relu~\citep{maas2013rectifier,Redmon2015YouOL,Ridniketal2021,GuoLiLer2024} tried tackling the dying \relu problem by allowing a small, non-zero gradient when the input is negative in order to keep the neurons active and the gradients flowing, reducing the risk of dead neurons.

\paragraph{Adaptive activation functions} 
\citet{Leeetal2022} propose the Adaptive Swish(\ash) activation, which uses stochastic sampling of the top-k percentile elements. It generalizes the \swish~\citep{Ramachandran2018SearchingFA}, which uses adaptive thresholding to select the values in the top percentiles and is set to zero otherwise. It is an example of a stochastic activation. %

\paragraph{Dropout and structured Dropout variants} In the original dropout paper~\citep{srivastava2014dropout}, the authors
propose a regularization technique to reduce overfitting and improve generalization of a neural network. It consists of setting to zero a subset of neurons at each training step. Consequently, the dropped neurons do not contribute to the forward pass or receive weight updates during back-propagation. At inference time, all neurons are used and their outputs are scaled by the dropout probability. 

LayerDrop~\citep{fan2019reducing}
randomly drops entire layers during training, hence, it encourages the model to be robust to missing layers. At inference time, some layers can be pruned, trading off speed and accuracy as needed. While the method does not make the model sparse in the usual sense, it induces structured sparsity in the computation graph during training.
Other works also introduce structured dropout variants such as DropBlock~\citep{ghiasi2018dropblockregularizationmethodconvolutional}, Bayesian dropout~\citep{GalGhah2016}, or Beit~\citep{Bao2021BEiTBP} and  masked-autoencoder~\citep{Heetal2022} in computer vision, among others.

\paragraph{Quantization approaches}

\citet{fan2020training} propose Quant-noise, that mimicks quantization during training by introducing noise to a random subset of weights for each forward pass enabling high compression ratios while maintaining the original model performance. It uses the Straight-Through estimator (STE)~\citep{Bengio2013EstimatingOP,hinton2012neural} to compute the gradients. 
This training technique ensures that the model is pretrained to observe both the train-time (unquantized) and the
inference-time (quantized) models. This
ensures proper optimization, bypassing the flat gradient caused by
quantization and reducing the discrepancy that results from the late quantization of the
model weights.

\paragraph{Sparsity by design} 
Some works propose to enable sparsity directly in the
architecture, for instance, the Mixture of Experts  (MoE) or the Product-Key Memory (PKM). The PKM architecture~\citep{lample2019large} uses a memory layer for neural networks which enables the model to access a large learnable memory and thus, it enables long term memory capabilities. It leverages product quantization (PQ)~\citep{JeDouSch2011} by splitting the key in two parts and using each part in separate codebooks. The combination of each PQ index enables the model to access a larger memory space efficiently. At each forward pass, only a small subset of the memory is accessed, making it computationally efficient.

Mixture of Expert (MoE) models ~\citep{Yang2024Qwen25TR,wei2024skyworkmoedeepdivetraining,deepseekv2, Jiang2024MixtralOE} dynamically select and activate the relevant subset of parameters based on the characteristics of the input data. The MoE approach allows MoE models to expand their capacity without proportionally increasing computational
complexity. See~\citet{MuLin2025} for an overview of the MoE and references therein.

\section{Using different activations at train and test time}
\label{sec:method}

This section introduces two strategies for improving the optimization during pretraining using an optimization-compliant activation, while preparing the model to a potentially different activation at test time. 
First we introduce the \swift fine-tuning approach. 
Then we introduce our Stochastic Activation \stocha. 

\subsection{Fine-tuning with \relu: \swift}

In the following, we use \silu and \relu as our training and inference activations. 
For reference, they are defined in $\mathbb{R}\rightarrow\mathbb{R}$ as: 
\begin{equation}
    \textrm{\relu}(x) = \max(x, 0)
    ~~~~~~~~~~
    \textrm{\silu}(x) = x\sigma(x),
\end{equation}
where $\sigma(x)=1/(1+\exp(-x))$ is the sigmoid function.
We choose these two activations because \silu is one of the best options in terms of accuracy, while \relu is simple and sparse. 
The two activations are also similar: same asymptotes at $-\infty$ and $+\infty$, and the same value at 0.
\silu is differentiable twice (unlike \relu) and, interestingly, non-monotonous.

In our proposed approach, the training operates as follows:
\begin{itemize}
    \item Most of the training steps (during a proportion $1-\alpha$ of the total number of iterations) are carried out with a first activation that is deemed preferable for training. We typically employ \silu for this stage. 
    \item We then switch the activation to that used for inference for the rest of the training. 
\end{itemize}

\begin{figure}
\begin{center}
\includegraphics[width=1\linewidth]{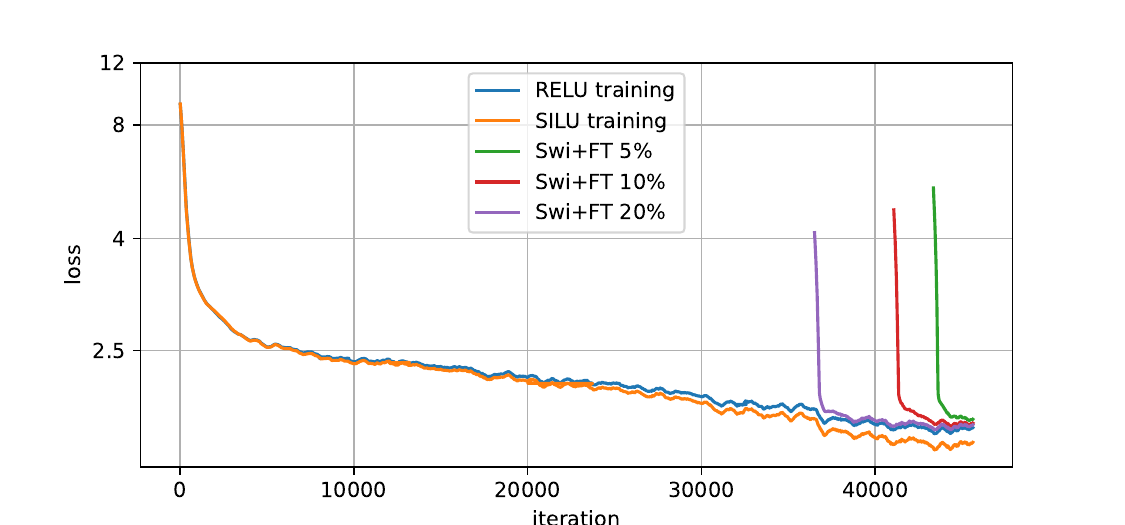}
\end{center}
\caption{
\swift: Training loss. Most of the training is carried out with \textcolor{orange}{\silu}, with $\alpha*100\%$=\textcolor{green!50!black}{5\%}, \textcolor{red}{10\%} and \textcolor{violet}{20\%} of the final steps using \textcolor{blue}{\relu}. 
Note the loss spike when we switch the activation. The model rapidly recovers and converges to a regime where \relu is performing well while providing sparsity. 
\label{fig:silu_finetuned_w_relu}
}
\end{figure}

We mostly set $\alpha=0.05$ or $\alpha=0.1$, which mean that only 5\% or 10\% of the training steps are carried out using the inference-time activation.  
We do not re-initialize the parameters of the optimizer when switching between activations, and similarly we do not use any warm-up. 
This does not disrupt the optimization because the \silu and \relu activations are relatively similar. We observe a spike in the loss at the time we change the activation, see Figure~\ref{fig:silu_finetuned_w_relu}. However, the optimization rapidly recovers. 
In practice, the fine-tuning replaces the last iterations of the pretraining. 
The learning rate follows a cosine schedule which gradually reduces it to 1/100\textsuperscript{th} of its peak value. 
Therefore, at 5\% or 10\% of the end of the training, the learning rate is already 60$\times$ or 29$\times$ lower than its peak, which is compatible with a fine-tuning regime. 

\subsection{Stochastic activation: \stocha} 
A stochastic function, parametrized by a random variable $\omega$, is a function 
\begin{equation}
    y=\Psi(x,\omega)
\end{equation} 
that maps inputs $x\in \mathbb{R}$ to output $y\in \mathbb{R}$ with randomness involved.  
The dependence on $\omega$ emphasizes that the outcome depends on an underlying probability space. 
In that sense, the function $\Psi(\cdot,\omega)$ is deterministic for each realization of $\omega$, but is stochastic overall.
In particular, we consider the case depicted in Figure~\ref{fig:bernoulli}, where $\omega \sim \mathrm{Bernoulli}(p)$ is a binary random variable parametrized by a parameter $p$: $\omega \in \{0, 1\}$ such that $\mathbb{P}(\omega=1)=p$ and $\mathbb{P}(\omega=0)=1-p$. 
In that case, the stochastic function $\Psi_p(\cdot)$ is defined such that
\begin{equation}
\textrm{if } x<0,       \ \Psi_p(x) = (1-\omega) \times \relu(x) + \omega \times \silu(x),
\end{equation}
which corresponds to randomly selecting between the \relu and \silu activations for $x<0$. 
If $x \geq 0$ we choose $\Psi_p(x) = x$ or $\Psi_p(x) = \silu(x)$, see the baselines description below.

This strategy ensures that the network is compatible with two regimes. The first one, drawn with probability $1-p$, is the inference-time mode, where we prepare the network to employ \relu during generation, in order to exhibit sparsity. 
The second mode aims to facilitate optimization during training. 
The choice of the \silu activation is motivated by the regular deterministic gated design by \citep{Shazeer2020GLUVI} adopted by most state-of-the-art LLMs.

\paragraph{Notation} 

To specify an activation, we separately define the function for the positive and negative range of inputs. 
    For example \texttt{R-S+} means that \relu is used for the negative range and \silu for the positive.
    When \stocha is used, we indicate \texttt{[S|R]-S+}, which means that for the negative range, we sample \silu with probability $p$ and \relu with probability $1-p$. 
\medskip
\begin{minipage}{0.66\linewidth}
\paragraph{Baselines}
The two natural baselines are the deterministic functions \silu and \relu. 
We also introduce two non-stochastic baselines in order to disentangle the effect that could come from combining \silu and \relu separately in the positive and negative domain: these baselines are denoted by \texttt{S-R+} and \texttt{S-R+}.
\end{minipage}
\hfill
\begin{minipage}{0.31\linewidth}
\begin{tabular}{lll}
\toprule
       & $x<0$               & $x\geq 0$             \\
\midrule
\relu  & $0$                 & $x$                   \\
\silu  & $x \cdot \sigma(x)$ & $x \cdot \sigma(x)$   \\
\rmsp  & $0$                 & $x \cdot \sigma(x)$   \\
\texttt{S-R+}  & $x \cdot \sigma(x)$ & $x$ \\
\bottomrule
\end{tabular}
\end{minipage}

\paragraph{Discussion}
The stochastic strategy resembles activation dropout~\citep{srivastava2014dropout}, which can be regarded as a particular case of our method where one of the activations is the null function. However, the objective of dropout is to avoid overfitting. 
Our motivation is closer to Quantization-aware training~\citep{Jacob2017QuantizationAT}, more specifically, to the QuantNoise strategy of~\citet{fan2020training}, where the model is pretrained to observe both the train-time (unquantized) and the inference-time (quantized) models. 
In QuantNoise, using these two modes during training time ensure both the proper optimization, without suffering the flat gradient inherent to quantization, while reducing the discrepancy that results from the late-quantization of the model weights.

\paragraph{Alternative construction of a stochastic activation.}
An alternative construction is to randomly select between the identity function $x \mapsto x$ and the constant zero function $x \mapsto 0$ with a sigmoidal probability $\sigma(x)$. As a result, in expectation this function is given by 
\begin{equation}
    \mathbb{E}[\mathrm{sa(x)}] = (1-\sigma(x)) \cdot 0 + \sigma(x) \cdot x = \sigma(x) \cdot x,  
\end{equation}
where we recognize the $\silu(x)$ function.
While the simplicity of this construction is mathematically appealing, our preliminary experiments revealed that it does not work very well.

\subsection{Inference-time strategies and evaluation }

At test time, we evaluate and analyze models trained with \swift and/or \stocha as follows:
\paragraph{\relu at test time} This is how we can enable sparsity. The corresponding evaluations therefore measure the performance on benchmarks when using this activation at test time. 

\paragraph{Exploiting sparsity}
On an input $x\in\mathbb{R}^D$, the gated FFN computes: 
\begin{equation}
    y = W_2 \times (\relu(W_1 \times x) \odot (W_3 \times x)) 
    \textrm{  with }
    W_1, W_3 \in \mathbb{R}^{N\times D}
    \textrm{  and }
    W_2 \in \mathbb{R}^{D\times N},
\label{eq:ffn}
\end{equation}
assuming column vectors and noting $\odot$ the element-wise multiplication. 
When the activation $\relu(W_1 \times x)$ has a fraction $s$ of zeros, the multiplications by $W_2$ and $W_3$ can exploit this sparsity: the baseline of $3ND$ FLOPS reduces to $(3 - 2s)ND$. 

Exploiting the sparsity to increase the computational throughput is not straightforward.  
At training time, the runtime is dominated by matrix-matrix multiplications, where even a 90\% sparsity rate does not yield efficiency gains.
At inference time with one prompt at a time, the bottleneck is the memory access used during matrix-vector multiplications. 
When $W_2$ is stored by rows and $W_3$ by columns, the sparsity can be exploited to avoid a fraction $s$ of the memory reads, that are contiguous~\citep{liu2023deja,liu2024training}. 
This implementation nearly yields the expected speedup (see experimental section).

\paragraph{Stochastic activation at test time} 

The following only applies to the \stocha strategy: we evaluate the performance by leveraging the randomness at test time, i.e., in this case, we do not use \relu. This choice is interesting for two reasons:
\begin{enumerate}
    \item To quantify the effect of the activation discrepancy between train and test. 
    \item To generate multiple outputs from the same prompt with the randomness of \stocha. 
\end{enumerate}

For the second usage, the standard way to generate multiple outputs from the same prompt is to replace the greedy decoding with a random sampling of the token from its probability distribution. 
This sampling can be tuned by setting a softmax temperature $T$ which adjusts between completely uniform sampling ($T\rightarrow\infty$) and strict maximum sampling ($T\rightarrow 0$). 
In both cases, we keep the one generated output with the highest normalized log likelihoods, \emph{i.e.}, the per-token average log-likelihood, as predicted by the model.

\section{Experiments with large language models}
\label{sec:experiments}

\subsection{Experimental setting}

\paragraph{Model architecture} We train dense decoder-only models. The transformer blocks use grouped-query attention~\citep{ainslie-etal-2023-gqa}. These models use RMSNorm~\citep{Zhang2019RootMS} with prenormalization, rotary positional positional encoding (RoPE)~\citep{Su2021RoFormerET} with $\theta = 500 000$ and train with document causal masking. We use the~\silu activation~\citep{Shazeer2020GLUVI} for the \silu baseline. The structure of our \amaiasmall and \amaiamed models is detailed in Table \ref{tab:architecture} in Appendix~\ref{sec:architecture}.

\paragraph{Training hyper-parameters} We train the models with AdamW optimizer~\citep{Loshchilov2017DecoupledWD} with $\beta_1$\,=\,$0.9$, $\beta_2$\,=\,$0.95$, learning rate of $lr$\,=\,$3\times10^{-3}$, weight decay of $0.1$, and gradient clipping at norm $1.0$. 
After 2000 steps of linear warm-up, 
we use a cosine decay learning rate schedule with peak learning rate  $8\times10^{-4}$ and decay 
by a factor of $1/100$ over the training horizon.

\paragraph{Tokenizer} We use the Llama3~\citep{dubey2024llama} tokenizer, which is a fast Byte-Pair Encoding tokenizer implemented with TikToken.2 The vocabulary contains 128 000 regular tokens as well as 256 reserved tokens.

\paragraph{Pre-training} We pre-train the \amaiasmall and \amaiamed models with  47B and 80B tokens, respectively, from a diverse collection of
mostly English natural language and coding data. We use a batch size of 1M tokens and a
context length of 8192 tokens.

\paragraph{Evaluation Benchmarks}

We employ two types of benchmarks for zero or few-shot evaluation, which we describe in more detail in Appendix~\ref{sec:benchmarks} and Table~\ref{tab:benchmarksconfigs}. The first type is code generation tasks: HumanEval+~\citep{evalplus} and  MBPP~\citep{chen2021codex}.  %
The second type consists of common sense and general reasoning: HellaSWAG\citep{Zellers2019HellaSwagCA}, ARC\citep{Clark2018ThinkYH}, PIQA~\citep{Bisk_Zellers_Lebras_Gao_Choi_2020}, OBQA~\citep{mihaylov-etal-2018-suit}, WinoGrande~\citep{Sakaguchi_LeBras_Bhagavatula_Choi_2020}, NQ~\citep{kwiatkowski-etal-2019-natural}, RACE~\citep{Lai2017RACELR}, TQA~\citep{TQA2017} and GSM8K~\citep{Cobbe2021TrainingVT}. 

\subsection{Performance analysis of \swift and \stocha with \relu at inference time}

\begin{table}
\centering
\resizebox{\textwidth}{!}{
{%
\begin{tabular}{l|ccccc|ccc|ccc}%
\toprule
& \multicolumn{4}{c}{Training activation} &   & \multicolumn{3}{c}{\amaiasmall}  &  \multicolumn{3}{c}{\amaiamed}\\
&\multicolumn{2}{c}{$x<0$}& $x>0$ & $p$ &         & train  & val &  val    & train  & val &  val         \\
\cmidrule(l){2-3}
\cmidrule(l){4-4} 
Activation     &  $p$ & $1-p$ &  &  & \swift  &  & \relu & \stocha &  & \relu & \stocha  \\
\midrule
\silu          &\multicolumn{2}{c}{\cSiLU} &\cSiLU&  - & \xmark & 2.105 & \addstar{2.122} &     & 1.966  & \addstar{1.974}  & \\
\relu          &\multicolumn{2}{c}{\cReLU} &\cReLU&  - & \xmark & 2.140 & \textcolor{blue}{2.161}         &     & 2.027  & 2.043  &  \\
\texttt{S-R+}  &\multicolumn{2}{c}{\cSiLU} &\cReLU&  - & \xmark & 2.101 & \addstar{2.124}                          &     & 1.970  & \addstar{1.980} &  \\ 
\texttt{R-S+}  &\multicolumn{2}{c}{\cReLU} &\cSiLU&  - & \xmark & 2.123 & \addstar{2.151}                          &     & 2.016  & \addstar{2.033}  & \\
\midrule   
\stosilu       &\cSiLU&\cReLU&\cReLU& 0.3  & \xmark & 2.120 & 2.363 & 2.146  & 1.993  & 2.257  & 2.006 \\
\stosilu       &\cSiLU&\cReLU&\cReLU& 0.5  & \xmark & 2.120 & 2.507 & 2.145  & 1.990  & 2.889  & 1.999 \\
\stodoublesilu &\cSiLU&\cReLU&\cSiLU& 0.3  & \xmark & 2.115 & 2.305 & 2.143  & 1.987  & 2.257  & 1.996 \\
\stodoublesilu &\cSiLU&\cReLU&\cSiLU& 0.5  & \xmark & 2.115 & 2.530 & 2.143  & 1.984  & 2.753  & 1.995 \\

\stosilu       &\cSiLU&\cReLU&\cReLU& 0.3  & \cmark & 2.123 & 2.141 & 2.251  & 1.988  & 1.998  & 2.177 \\
\stosilu       &\cSiLU&\cReLU&\cReLU& 0.5  & \cmark & 2.129 & 2.148 & 2.307  & 1.989  & 2.002  & 2.306 \\
\stodoublesilu &\cSiLU&\cReLU&\cSiLU& 0.3  & \cmark & 2.120 & 2.138 & 2.221  & 1.982  & 1.992  & 2.103 \\
\stodoublesilu &\cSiLU&\cReLU&\cSiLU& 0.5  & \cmark & 2.125 & 2.144 & 2.301  & 1.985  & 1.994  & 2.234 \\
\bottomrule
\end{tabular}
}
}
\caption{
The train loss is computed over the last 500 steps of the training of~\amaiasmall, the val loss is measured after training, on a different set of text and code, using the \relu activation$^\star$ or \stocha, i.e. the same activation used at train time (possibly deterministic).   
If \swift is enabled, we switch to \relu for the last $5\%$ steps. 
{$^\star$: \footnotesize for the deterministic baselines \silu, \texttt{S-R+} and \texttt{S+R-}, we do the inference with the same activation used at train-time (not \relu). }
}
\label{tab:nll}
\end{table}

In this section we analyze the effect of our proposal when using \relu at test time. 
In Appendix~\ref{sec:sparsity}, we provide a complementary analysis of the sparsity. Depending on the setting, the average rate of 0s can be higher than 90\%, when using the \relu at test time.

\paragraph{Cross-entropy performance}
Table \ref{tab:nll} summarizes the impact on the training and validation losses of multiple choices with the \amaiasmall{} and ~\amaiamed{}  models. 
We observe that the training loss using stochastic activation at train time is lower than that obtained with \relu. %
However the validation entropy is not competitive per se, due to the remaining train-inference discrepancy of activation. 
This is solved by \swift: switching to the \relu activation function and fine-tuning for the last 5\% or 10\% steps of the training steps drastically boosts the test-time inference. These results outperform the results obtained with regular \relu training, while using the same activation at test time. 

\paragraph{Fast inference with \relu sparsity}

\begin{figure}
\centering
\begin{tabular}{cc}
CPU (one core of Xeon 8462Y+) & GPU (H100 80GB HBM3)\\
\includegraphics[width=0.45\linewidth]{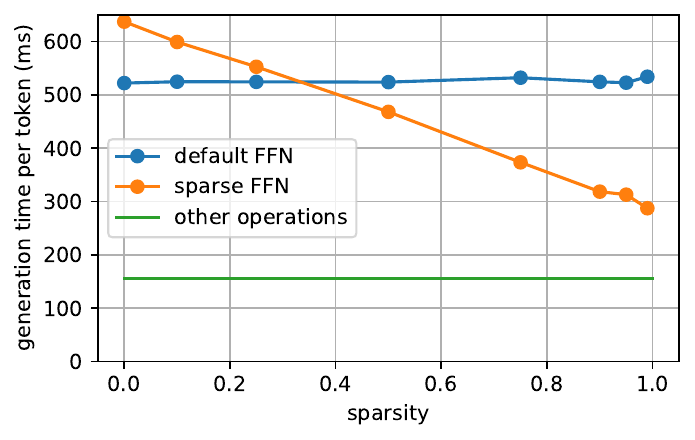} & 
\includegraphics[width=0.45\linewidth]{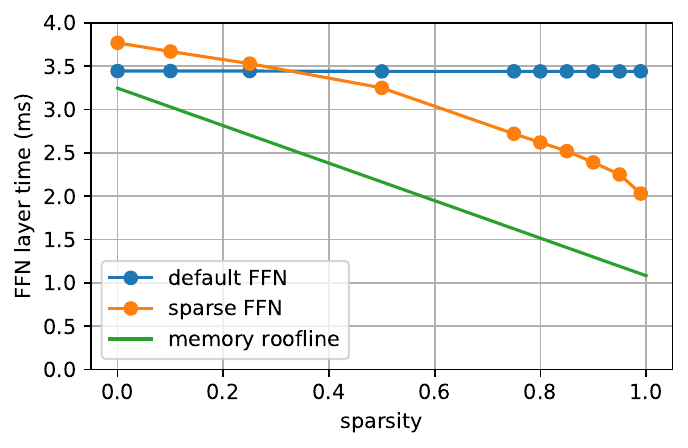} \\
\end{tabular}
    \caption{
Inference times for 1 token, as a function of the activation sparsity, with a \amaiamed model trained with \swift.
For the CPU  we indicate the total inference time, including ``other operations": the attention layers (these are not dominant because the generation is limited to 200 tokens), the normalization and the execution overheads.
At 90\% sparsity the speedup is $\times$1.65.
For the GPU, we measure only the FFN inference time.
Note that the impact on inference is less  important because at this speed and model scale the overheads are dominant. 
We also indicate the minimum runtime that could be obtained given the GPU's memory bandwidth (roofline model). 
    \label{fig:runtimes}
    }
\end{figure}

The activation sparsity can be exploited to avoid fetching 90\% of the matrices $W_1$ and $W_3$ of the FFN (see Eq.~\ref{eq:ffn} and detailed sparsity rates in Appendix~\ref{sec:sparsity}). 
As a consequence, we observe a direct benefit on CPU: Figure~\ref{fig:runtimes} shows that 90\% sparsity directly translates into a 65\% speedup. 
On GPU, the speedup of FFN inference is around 1.5$\times$. 
The GPU implementation requires different blocking and tiling schemes depending on the sparsity.
However, because the overall inference is much faster, different overheads dominate raw computation time of the FFN layers. 
These include the KV-cache management, Python interpretation, attention masks and Rope management, etc. 
We expect these overheads to be less significant on larger models or carefully optimized software stacks.

\paragraph{Absence of dead neurons with \swift}
Figure~\ref{fig:cdfs}(\textit{left}) shows a sudden jump in the \relu CDF, which indicates there are many activations that are exactly zero \emph{on input} of \relu. 
The \relu plot in Figure~\ref{fig:cdfs}(\textit{right}) shows that a large fraction of the rows of the linear layer $W_1$ (Eq~\ref{eq:ffn}) are near zero (\ie their norm is less than 1/1000 the average row norm of the matrix). 
This means that these rows are unused: they are ``dead neurons''. 
In contrast, the \swift plot in Figure~\ref{fig:cdfs}(\textit{right}) we observe that using \silu at pre-training prevents these dead neurons.
This explains why our approach obtains significantly better results with \relu than the vanilla training of the model using the \relu from scratch. 

\begin{figure}[b]
\begin{minipage}{0.65\linewidth}
\includegraphics[width=\linewidth]{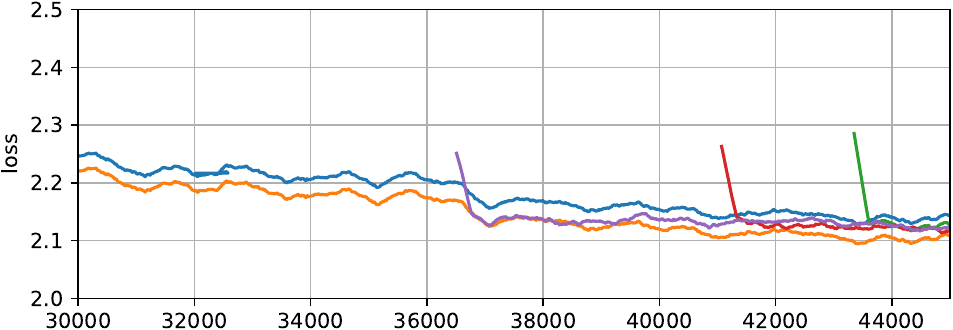}
\end{minipage}
\hfill
\begin{minipage}{0.3\linewidth}
{\footnotesize
    \begin{tabular}{l|cc}
    \toprule
                         &  \multicolumn{2}{c}{final loss}\\
    \phantom{0}$\alpha$  &  train & val (\relu) \\
    \midrule
    0       &  2.115   &  2.530    \\
    0.05    &  2.125   &  2.144    \\
    0.10    &  2.119   &  2.145    \\
    0.20    &  2.126   &  2.155    \\
    \bottomrule
    \end{tabular}}
\end{minipage}
\caption{
Training loss with \swift and \stocha: \stodoublesilu activation with $p=0.5$ for $\alpha*100\%$=\textcolor{RedViolet!50!black}{$5\%$}, \textcolor{red!75!black}{$10\%$} and \textcolor{green!50!black}{$20\%$}, relative to \textcolor{Blue!80}{\relu} and \textcolor{orange!80!black}{\silu}. This shows that the \swift strategy needs to be combined with \stocha to provide good models operating with \relu compared to finetuning \silu with \relu alone (Figure~\ref{fig:silu_finetuned_w_relu}).
This plot this is zoomed in relative to Figure~\ref{fig:silu_finetuned_w_relu}.
\label{fig:loss_stocha+swift}
}
\end{figure}

\paragraph{Complementarity between \stocha and \swift when fine-tuned with \relu}
Figure~\ref{fig:loss_stocha+swift} shows the training loss when using~\stocha and \swift jointly. When switching the activation to \relu, we observe a spike in the loss, but the optimization rapidly recovers, and  converges to a model that has better performance than the one trained with \relu from scratch. 
In contrast, in the case where we employ \swift alone (Figure~\ref{fig:silu_finetuned_w_relu}), fine-tuning with~\relu after pretraining with \silu is not enough to obtain performance improvements. 
As a consequence, the \stocha and \swift approaches are complementary. 

\paragraph{Impact of fine-tuning with \relu for the last $\alpha*100\%$ steps} In Figure~\ref{fig:loss_stocha+swift}, we compare the final loss reported for different values of $\alpha$. In that case, setting $\alpha=0.1$ is a good trade-off since the corresponding loss is lower than the \relu loss. 
\paragraph{Continuous pre-training with \stocha and \swift} Starting from a pre-trained \amaiasmall model with \silu, we can leverage either \stocha or \swift  methods for continuous pretraining (CPT), see Appendix~\ref{sec:cpt} for experimental details. 
In such case, the best strategy is to use \swift, i.e., simply fine-tune with RELU the model initially trained with \silu to obtain the best trade-off between sparsity and loss (Table~\ref{tab:cpt}). This pattern is the opposite of what we observe for LMs trained from scratch, where the \stocha and \stocha+\swift offer the best trade-off between sparsity and loss (Table~\ref{tab:nll}). 

\paragraph{Generalizability of \stocha and \swift for any pair of activations} In Appendix~\ref{sec:tanhrelu}, we explain how our methods can be generalized for any pair of activation functions and include an illustrative example with the (\tanh,\relu) activations.

\subsection{Performance on downstream tasks}

\paragraph{Detailed results per benchmark} 
Table~\ref{tab:benchmark_detail} reports the results for standard code generation, common sense and general reasoning benchmarks detailed in Appendix~\ref{sec:benchmarks}. We consider multiple \stocha models using few-shot or zero shot prompting, see Table~\ref{tab:benchmarksconfigs} in Appendix~\ref{sec:benchmarks} for more details. 
The model with \silu (topline) is significantly better that a regular model with \relu. However, our models trained with \stocha are slightly better or on par with \silu: either the model fine-tuned with \swift and using \relu at inference time, or even the model that uses \stocha at test time.

\begin{table}[t]
\resizebox{\textwidth}{!}{
\small
\begin{tabular}{l@{}cccc|ccccc}
\toprule
$\downarrow$ \textbf{Benchmark/metric } &  \multicolumn{4}{@{}c@{\ }}{\amaiasmall}  &  \multicolumn{4}{@{}c@{\ }}{\amaiamed}  \\
\cmidrule(l){2-4}
\cmidrule(l){5-9} 
 & \multicolumn{2}{@{}c@{\ }}{(a) baselines}  & (b)\,\swift & (c)\,\stocha & \multicolumn{2}{@{}c@{\ }}{(a) baselines}  & (b)\,\swift & (c)\,\stocha  \\ 
~\hfill train activation $\rightarrow$     &  \silu & \relu  &\stodoublesilu& \stodoublesilu &  \silu & \relu  &\stodoublesilu& \stodoublesilu \\
~\hfill inference activation $\rightarrow$ & \silu & \relu & \relu & \stodoublesilu & \silu & \relu & \relu & \stodoublesilu \\
\midrule
hellaswag/acc\_char          & 0.585 & 0.561 & 0.574  & 0.576  & 0.684 & 0.633 & 0.671 & 0.678 \\
winogrande/acc\_char         & 0.593 & 0.571 & 0.568  & 0.568  & 0.657 & 0.615 & 0.630 & 0.620 \\
arc\_easy/acc\_char          & 0.568 & 0.562 & 0.600  & 0.562  & 0.675 & 0.642 & 0.679 & 0.671\\
arc\_challenge/acc\_char     & 0.313 & 0.286 & 0.331  & 0.314  & 0.390 & 0.348 & 0.396 & 0.376 \\
piqa/acc\_char               & 0.732 & 0.720 & 0.724  & 0.720  & 0.767 & 0.751 & 0.765 & 0.761 \\
obqa/acc\_char               & 0.346 & 0.340 & 0.378  & 0.340  & 0.390 & 0.380 & 0.384 & 0.408 \\
race.middle/acc\_char        & 0.518 & 0.516 & 0.509  & 0.498  & 0.565 & 0.538 & 0.559 & 0.549\\
race.high/acc\_char          & 0.382 & 0.379 & 0.372  & 0.375  & 0.414 & 0.402 & 0.407 & 0.416  \\
human\_eval\_plus/pass@1     & 0.073 & 0.067 & 0.049  & 0.055  & 0.128 & 0.110 & 0.128 & 0.116 \\
mbpp/compiles@1              & 0.978 & 0.970 & 0.960  & 0.980  & 0.992 & 0.980 & 0.990 & 0.982 \\
tqa/f1                       & 0.243 & 0.217 & 0.229  & 0.232  & 0.351 & 0.293 & 0.327 & 0.342 \\
nq/f1                        & 0.123 & 0.107 & 0.121  & 0.113  & 0.169 & 0.146 & 0.170 & 0.145\\
\midrule
average performance          & \textbf{0.454} & 0.441 & \underline{0.451}  & 0.444  & \textbf{0.515} & 0.486 & \underline{0.509} & 0.505 \\
\bottomrule
\end{tabular}}
\vspace{-0.5em}
\caption{Performance per benchmark of (a) \relu and \silu baselines for \amaiasmall and \amaiamed compared to (b) models with \stocha+\swift at train time and \relu at test time, and (c) models with \stocha at train and test time. We use the model with the best perplexity on val namely $p=0.3, \alpha=0.05$ for \swift and $p=0.5$ for \stocha. The average performance for the topline (\silu) is in bold, second best (\stocha+\swift) is underlined.
\label{tab:benchmark_detail}
\label{tab:benchmark_detail3B}
}
\end{table}

\begin{figure}
\vspace{-1em}
\includegraphics[height=6.5cm]{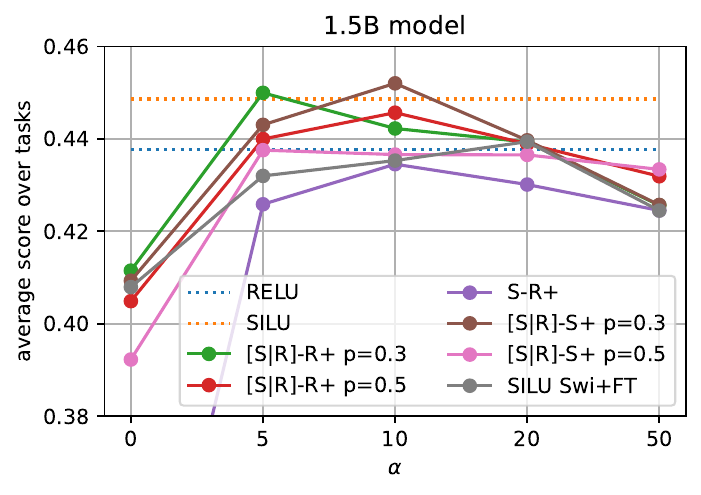}%
\hfill
\includegraphics[height=6.5cm]{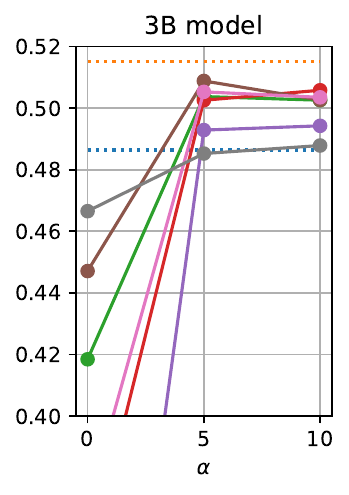}
\vspace{-1.0em}
\caption{\textbf{\swift: analysis of the fine-tuning rate~$\alpha$.} 
Average performance over the benchmarks as a function of the percentage $\alpha$ of steps for which we switch to the \relu activation at the end of training. 
We use \relu at inference time. 
\label{fig:amaiasmallevals}}
\end{figure}

\paragraph{Performance when varying $\boldsymbol{\alpha}$ with \swift} 
Figure~\ref{fig:amaiasmallevals} shows that we can slightly surpass the \silu baseline average performance if we first use a stochastic activation function during the ~\amaiasmall{} model training and then switch to the \relu activation for the last $\alpha*100\%$ of the training steps, for $\alpha \in \{0.05,0.1,0.2\}$. 
The best performance is obtained with $\alpha=0.05$ or $\alpha=0.1$ for the \amaiasmall{} model.

\subsection{Exploiting \stocha at test time}

\paragraph{Effectiveness of \stocha a test time}
In Table \ref{tab:nll}, in addition to the results with \relu at test time, we also report the train and validation losses obtained when employing \stocha at test time. We observe that (1) using stochastic activations for inference works surprisingly well in spite of the randomness. The results are between \relu and \silu in most configurations; (2) When using \stocha at test time, there is no need to fine-tune the model with \swift. This is expected since this strategy is intended to decrease the discrepancy with the test-time activation choice. 
Table~\ref{tab:varyingp} shows that the average benchmark performance generally increases when the stochastic mix approaches \silu. 
Therefore, \stocha is primarily useful as a way to generate multiple outputs for the same prompt.

\begin{table}
\centering
{\small
\begin{tabular}{l@{\quad}l@{\quad}|@{\quad}rr}
\toprule 
  $p$                  & \amaiasmall & \amaiamed \\
\midrule
0\phantom{.1} 
(\texttt{R-S+})     & 0.215  & 0.222 \\
0.3                 & 0.443  & 0.504 \\
0.5                 & 0.426  & 0.505 \\
0.7                 &  0.453 & 0.495\\
1.0 (\texttt{SILU}) & 0.454 & 0.515 \\
\bottomrule
\end{tabular}}
\caption{
    \stocha: Impact on benchmarks performance (avg) as a function of the \stocha $p$ for \stodoublesilu used at test time. 
The case $p=0$ corresponds to \texttt{R-S+} while $p=1$ corresponds to the baseline \silu. 
The performance increases with more \silu in the mix. 
However, the stochasticity can be used to increase the generation diversity. 
}
\label{tab:varyingp}
\end{table}

\begin{figure}[t!]
\vspace{-1em}
\centering
\resizebox{\textwidth}{!}{
\begin{tabular}{cc}
(a) F1 scores: & \\
NQ & TQA \\
\includegraphics[width=0.45\linewidth]{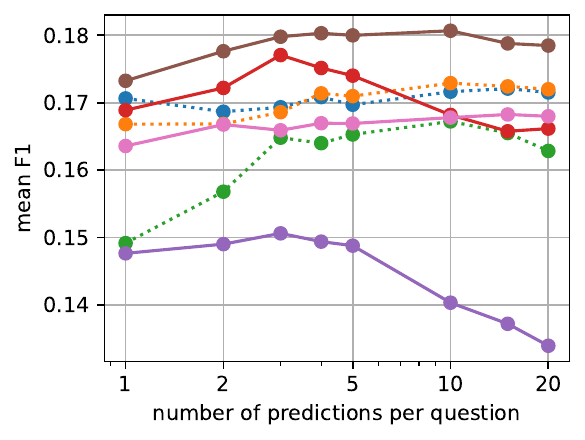} & 
\includegraphics[width=0.45\linewidth]{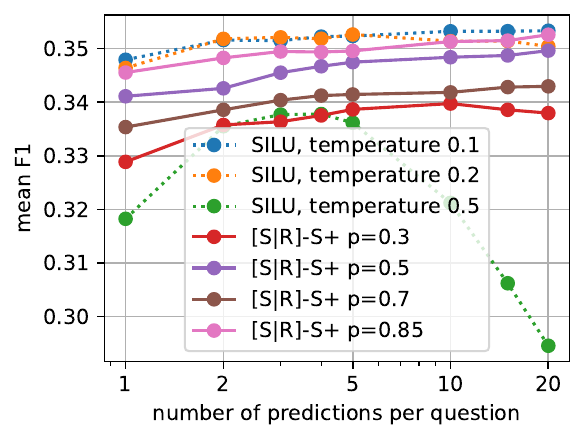} \\
(b) Average type-token ratio: &\\
\includegraphics[width=0.5\textwidth]{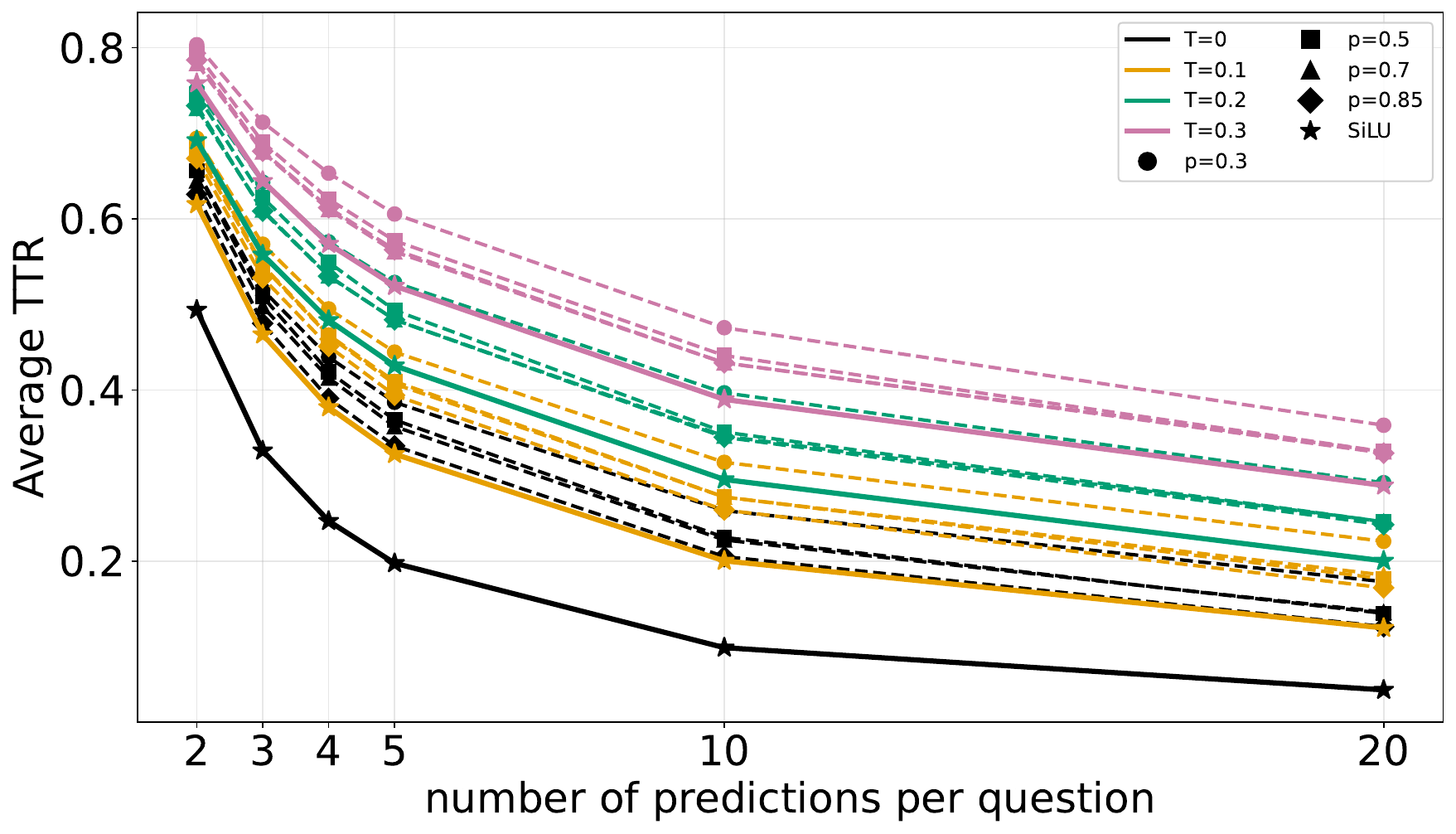} & 
\includegraphics[width=0.5\textwidth]{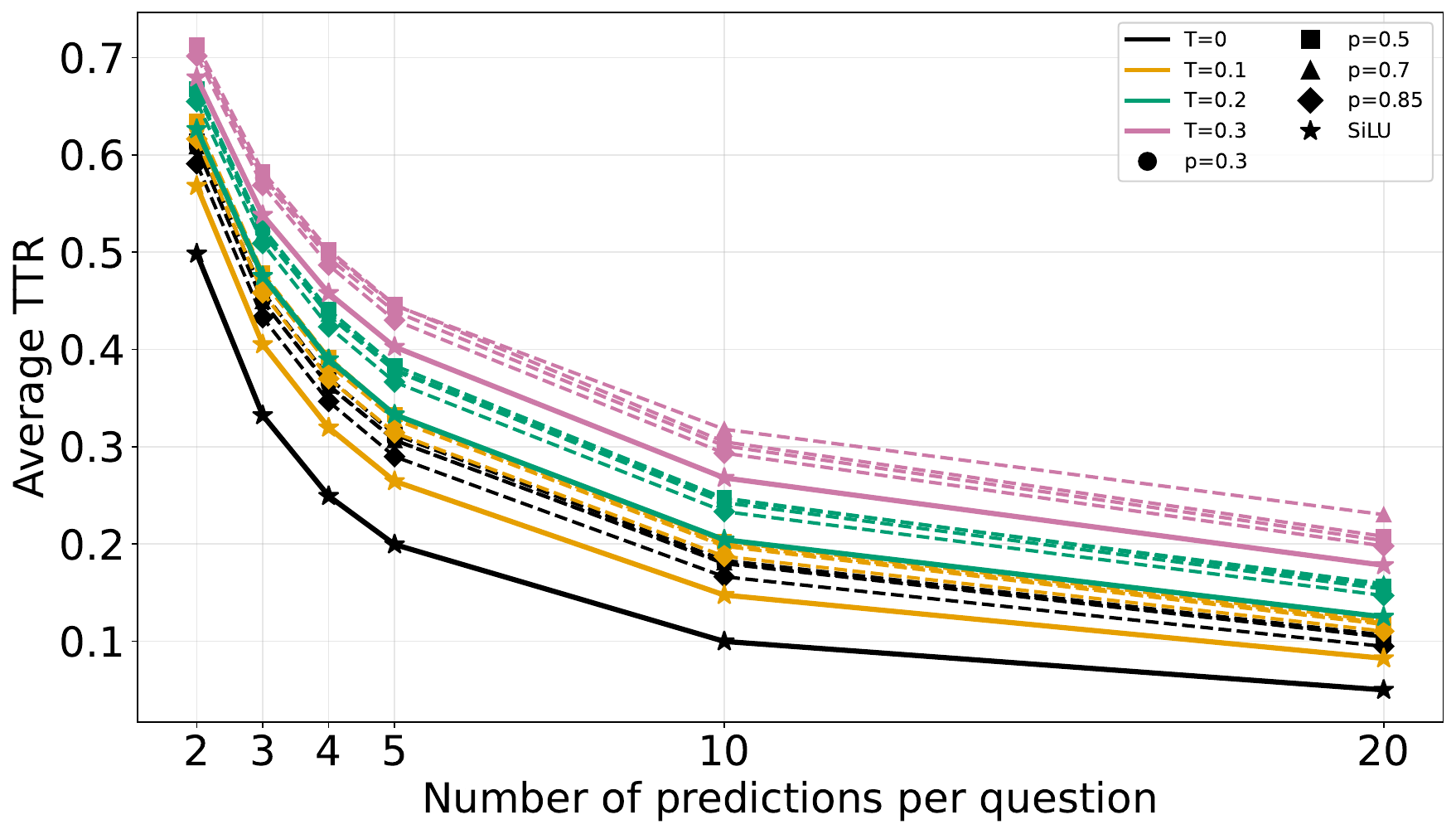} \\
\end{tabular}}
\vspace{-1em}
\caption{
    Comparison of diverse response generation methods in Q\&A benchmarks.
    (a) F1 scores when varying the temperature with \silu or \stodoublesilu $p$ with temp$=0$, the generations are scored by normalized log likelihoods.
    (b) TTR when varying the temperature in temperature sampling with \silu (solid lines) vs \stodoublesilu for different values of $p$ (dashed lines).
}
\label{fig:mutiplegen}
\end{figure}

\paragraph{Diversity of generated sequences ablation} 
Figure~\ref{fig:mutiplegen} shows that leveraging the \stocha activations stochasticity (combined with standard temperature sampling) generates more diverse sequences compared to using \silu with standard temperature sampling. See Appendix~\ref{sec:multigen_setup} for the experimental setup and Appendix~\ref{sec:multiple_generation} for qualitative examples of such generations. 
 
In Figure~\ref{fig:mutiplegen} (a), the curves are increasing, which means that (1) we obtain diverse generations and (2) that the normalized log-likelihood is a suitable scoring function. Specifically, for the NQ task, the \stocha activation with $p=0.7$ and temperature zero yields consistently higher performance than standard temperature sampling with the \silu activation (for temperatures in $\{0.1,0.2,0.5\}$). The \stocha activation with $p=0.3$ is also competitive for the number of generations up to five. In contrast, the results are subpar for the TQA task: standard temperature sampling with \silu activation yields higher performances for all number of generations for temperatures in $\{0.1,0.2\}$.

In Figure~\ref{fig:mutiplegen} (b), we observe that all the models that use 
\stocha activation functions have consistently higher average TTR for both tasks, for each temperature in $\{0,0.1,0.2,0.3\}$ compared to standard temperature sampling with \silu. This finding further confirms that the text diversity of generations is much better for these models.

\paragraph{Token-level diversity ablation} In Appendix~\ref{sec:multiforward}, we highlight that using \stocha at inference time is equivalent to applying Monte Carlo dropout~\citep{GalGhah2016} on the negative side of the input after activation. We then leverage this token level diversity at inference time by doing multiple forward passes per generated token using greedy decoding. As observed in the sequence level ablation, more diverse tokens are generated. However, we did not observe that this resulted in better performance if the maximum probability or majority vote selection strategies were used.

\section{Conclusion}
\label{sec:conclusion}

This paper introduces a novel stochastic activation that preserves the performance of a non-sparse activation, such as \silu, while better adjusting to the behavior of a sparse one, such as \relu, at test time. This improves the inference times for the FFN layers of a transformer, translating into a speedup of typically $\times1.65$ for the FFN processing on CPUs while almost preserving the accuracy of the non-sparse \silu activation.  
Finally, we explore how stochastic activations can be leveraged at test time to improve diversity in model generations.

\bibliography{main}

\begin{thebibliography}{52}
\providecommand{\natexlab}[1]{#1}
\providecommand{\url}[1]{\texttt{#1}}
\expandafter\ifx\csname urlstyle\endcsname\relax
  \providecommand{\doi}[1]{doi: #1}\else
  \providecommand{\doi}{doi: \begingroup \urlstyle{rm}\Url}\fi

\bibitem[Ainslie et~al.(2023)Ainslie, Lee-Thorp, de~Jong, Zemlyanskiy, Lebron,
  and Sanghai]{ainslie-etal-2023-gqa}
Joshua Ainslie, James Lee-Thorp, Michiel de~Jong, Yury Zemlyanskiy, Federico
  Lebron, and Sumit Sanghai.
\newblock {GQA}: Training generalized multi-query transformer models from
  multi-head checkpoints.
\newblock In Houda Bouamor, Juan Pino, and Kalika Bali (eds.),
  \emph{Proceedings of the 2023 Conference on Empirical Methods in Natural
  Language Processing}, pp.\  4895--4901, Singapore, December 2023. Association
  for Computational Linguistics.
\newblock \doi{10.18653/v1/2023.emnlp-main.298}.
\newblock URL \url{https://aclanthology.org/2023.emnlp-main.298/}.

\bibitem[Austin et~al.(2021)Austin, Odena, Nye, Bosma, Michalewski, Dohan,
  Jiang, Cai, Terry, Le, and Sutton]{Austin2021ProgramSW}
Jacob Austin, Augustus Odena, Maxwell Nye, Maarten Bosma, Henryk Michalewski,
  David Dohan, Ellen Jiang, Carrie~J. Cai, Michael Terry, Quoc~V. Le, and
  Charles Sutton.
\newblock Program synthesis with large language models.
\newblock \emph{ArXiv}, abs/2108.07732, 2021.
\newblock URL \url{https://api.semanticscholar.org/CorpusID:237142385}.

\bibitem[Bao et~al.(2021)Bao, Dong, and Wei]{Bao2021BEiTBP}
Hangbo Bao, Li~Dong, and Furu Wei.
\newblock Beit: Bert pre-training of image transformers.
\newblock \emph{ArXiv}, abs/2106.08254, 2021.
\newblock URL \url{https://api.semanticscholar.org/CorpusID:235436185}.

\bibitem[Bengio et~al.(2013)Bengio, L{\'e}onard, and
  Courville]{Bengio2013EstimatingOP}
Yoshua Bengio, Nicholas L{\'e}onard, and Aaron~C. Courville.
\newblock Estimating or propagating gradients through stochastic neurons for
  conditional computation.
\newblock \emph{ArXiv}, abs/1308.3432, 2013.
\newblock URL \url{https://api.semanticscholar.org/CorpusID:18406556}.

\bibitem[Bisk et~al.(2020)Bisk, Zellers, Le~bras, Gao, and
  Choi]{Bisk_Zellers_Lebras_Gao_Choi_2020}
Yonatan Bisk, Rowan Zellers, Ronan Le~bras, Jianfeng Gao, and Yejin Choi.
\newblock Piqa: Reasoning about physical commonsense in natural language.
\newblock \emph{Proceedings of the AAAI Conference on Artificial Intelligence},
  34\penalty0 (05):\penalty0 7432--7439, Apr. 2020.
\newblock \doi{10.1609/aaai.v34i05.6239}.
\newblock URL \url{https://ojs.aaai.org/index.php/AAAI/article/view/6239}.

\bibitem[Brown et~al.(2020)Brown, Mann, Ryder, Subbiah, Kaplan, Dhariwal,
  Neelakantan, Shyam, Sastry, Askell, et~al.]{brown2020language}
Tom~B Brown, Benjamin Mann, Nick Ryder, Melanie Subbiah, Jared Kaplan, Prafulla
  Dhariwal, Arvind Neelakantan, Pranav Shyam, Girish Sastry, Amanda Askell,
  et~al.
\newblock Language models are few-shot learners.
\newblock \emph{arXiv preprint arXiv:2005.14165}, 2020.

\bibitem[Chen et~al.(2021)Chen, Tworek, Jun, Yuan, de~Oliveira~Pinto, Kaplan,
  and team]{chen2021codex}
Mark Chen, Jerry Tworek, Heewoo Jun, Qiming Yuan, Henrique~Ponde
  de~Oliveira~Pinto, Jared Kaplan, and OpenAI team.
\newblock Evaluating large language models trained on code.
\newblock 2021.

\bibitem[Chowdhery et~al.(2022)Chowdhery, Narang, Devlin, and
  team]{chowdhery2022palmscalinglanguagemodeling}
Aakanksha Chowdhery, Sharan Narang, Jacob Devlin, and Google team.
\newblock Palm: Scaling language modeling with pathways, 2022.
\newblock URL \url{https://arxiv.org/abs/2204.02311}.

\bibitem[Clark et~al.(2018)Clark, Cowhey, Etzioni, Khot, Sabharwal, Schoenick,
  and Tafjord]{Clark2018ThinkYH}
Peter Clark, Isaac Cowhey, Oren Etzioni, Tushar Khot, Ashish Sabharwal, Carissa
  Schoenick, and Oyvind Tafjord.
\newblock Think you have solved question answering? try arc, the ai2 reasoning
  challenge.
\newblock \emph{ArXiv}, abs/1803.05457, 2018.
\newblock URL \url{https://api.semanticscholar.org/CorpusID:3922816}.

\bibitem[Cobbe et~al.(2021)Cobbe, Kosaraju, Bavarian, Chen, Jun, Kaiser,
  Plappert, Tworek, Hilton, Nakano, Hesse, and Schulman]{Cobbe2021TrainingVT}
Karl Cobbe, Vineet Kosaraju, Mo~Bavarian, Mark Chen, Heewoo Jun, Lukasz Kaiser,
  Matthias Plappert, Jerry Tworek, Jacob Hilton, Reiichiro Nakano, Christopher
  Hesse, and John Schulman.
\newblock Training verifiers to solve math word problems.
\newblock \emph{ArXiv}, abs/2110.14168, 2021.
\newblock URL \url{https://api.semanticscholar.org/CorpusID:239998651}.

\bibitem[DeepSeek-AI et~al.(2024)DeepSeek-AI, Liu, Feng, Wang, Wang, Liu, Zhao,
  Dengr, Ruan, Dai, Guo, et~al.]{deepseekv2}
DeepSeek-AI, Aixin Liu, Bei Feng, Bin Wang, Bingxuan Wang, Bo~Liu, Chenggang
  Zhao, Chengqi Dengr, Chong Ruan, Damai Dai, Daya Guo, et~al.
\newblock Deepseek-v2: A strong, economical, and efficient mixture-of-experts
  language model, 2024.
\newblock URL \url{https://arxiv.org/abs/2405.04434}.

\bibitem[Devlin et~al.(2019)Devlin, Chang, Lee, and
  Toutanova]{devlin2019bertpretrainingdeepbidirectional}
Jacob Devlin, Ming-Wei Chang, Kenton Lee, and Kristina Toutanova.
\newblock Bert: Pre-training of deep bidirectional transformers for language
  understanding, 2019.
\newblock URL \url{https://arxiv.org/abs/1810.04805}.

\bibitem[Dubey et~al.(2024)Dubey, Jauhri, Pandey, Kadian, Al-Dahle, Letman,
  Mathur, Schelten, Yang, Fan, et~al.]{dubey2024llama}
Abhimanyu Dubey, Abhinav Jauhri, Abhinav Pandey, Abhishek Kadian, Ahmad
  Al-Dahle, Aiesha Letman, Akhil Mathur, Alan Schelten, Amy Yang, Angela Fan,
  et~al.
\newblock The llama 3 herd of models, 2024.
\newblock URL \url{https://arxiv.org/abs/2407.21783}.

\bibitem[Fan et~al.(2019)Fan, Grave, and Joulin]{fan2019reducing}
Angela Fan, Edouard Grave, and Armand Joulin.
\newblock Reducing transformer depth on demand with structured dropout.
\newblock \emph{arXiv preprint arXiv:1909.11556}, 2019.

\bibitem[Fan et~al.(2020)Fan, Stock, Graham, Grave, Gribonval, Jegou, and
  Joulin]{fan2020training}
Angela Fan, Pierre Stock, Benjamin Graham, Edouard Grave, R{\'e}mi Gribonval,
  Herve Jegou, and Armand Joulin.
\newblock Training with quantization noise for extreme model compression.
\newblock \emph{arXiv preprint arXiv:2004.07320}, 2020.

\bibitem[Gal \& Ghahramani(2016)Gal and Ghahramani]{GalGhah2016}
Yarin Gal and Zoubin Ghahramani.
\newblock Dropout as a bayesian approximation: representing model uncertainty
  in deep learning.
\newblock In \emph{Proceedings of the 33rd International Conference on
  International Conference on Machine Learning - Volume 48}, ICML'16, pp.\
  1050–1059. JMLR.org, 2016.

\bibitem[Ghiasi et~al.(2018)Ghiasi, Lin, and
  Le]{ghiasi2018dropblockregularizationmethodconvolutional}
Golnaz Ghiasi, Tsung-Yi Lin, and Quoc~V. Le.
\newblock Dropblock: A regularization method for convolutional networks, 2018.
\newblock URL \url{https://arxiv.org/abs/1810.12890}.

\bibitem[Glorot et~al.(2011)Glorot, Bordes, and Bengio]{glorot2011deep}
Xavier Glorot, Antoine Bordes, and Yoshua Bengio.
\newblock Deep sparse rectifier neural networks.
\newblock In \emph{Proceedings of the fourteenth international conference on
  artificial intelligence and statistics}, pp.\  315--323. JMLR Workshop and
  Conference Proceedings, 2011.

\bibitem[Guo et~al.(2024)Guo, Li, and Lerman]{GuoLiLer2024}
Yinglong Guo, Shaohan Li, and Gilad Lerman.
\newblock The effect of leaky relus on the training and generalization of
  overparameterized networks.
\newblock \emph{CoRR}, abs/2402.11942, 2024.
\newblock \doi{10.48550/ARXIV.2402.11942}.
\newblock URL \url{https://doi.org/10.48550/arXiv.2402.11942}.

\bibitem[He et~al.(2022)He, Chen, Xie, Li, Dollár, and Girshick]{Heetal2022}
Kaiming He, Xinlei Chen, Saining Xie, Yanghao Li, Piotr Dollár, and Ross
  Girshick.
\newblock Masked autoencoders are scalable vision learners.
\newblock In \emph{2022 IEEE/CVF Conference on Computer Vision and Pattern
  Recognition (CVPR)}, pp.\  15979--15988, 2022.
\newblock \doi{10.1109/CVPR52688.2022.01553}.

\bibitem[Hendrycks \& Gimpel(2016)Hendrycks and
  Gimpel]{Hendrycks2016GaussianEL}
Dan Hendrycks and Kevin Gimpel.
\newblock Gaussian error linear units (gelus).
\newblock \emph{arXiv: Learning}, 2016.
\newblock URL \url{https://api.semanticscholar.org/CorpusID:125617073}.

\bibitem[Hinton(2012)]{hinton2012neural}
Geoffrey Hinton.
\newblock Neural networks for machine learning.
\newblock Coursera, video lectures, 2012.
\newblock Online course.

\bibitem[Jacob et~al.(2017)Jacob, Kligys, Chen, Zhu, Tang, Howard, Adam, and
  Kalenichenko]{Jacob2017QuantizationAT}
Benoit Jacob, Skirmantas Kligys, Bo~Chen, Menglong Zhu, Matthew Tang, Andrew~G.
  Howard, Hartwig Adam, and Dmitry Kalenichenko.
\newblock Quantization and training of neural networks for efficient
  integer-arithmetic-only inference.
\newblock \emph{2018 IEEE/CVF Conference on Computer Vision and Pattern
  Recognition}, pp.\  2704--2713, 2017.
\newblock URL \url{https://api.semanticscholar.org/CorpusID:39867659}.

\bibitem[Jiang et~al.(2024)Jiang, Sablayrolles, Roux, Mensch, Savary, Bamford,
  Chaplot, de~Las~Casas, Hanna, Bressand, Lengyel, Bour, Lample, Lavaud,
  Saulnier, Lachaux, Stock, Subramanian, Yang, Antoniak, Scao, Gervet, Lavril,
  Wang, Lacroix, and Sayed]{Jiang2024MixtralOE}
Albert~Q. Jiang, Alexandre Sablayrolles, Antoine Roux, Arthur Mensch, Blanche
  Savary, Chris Bamford, Devendra~Singh Chaplot, Diego de~Las~Casas, Emma~Bou
  Hanna, Florian Bressand, Gianna Lengyel, Guillaume Bour, Guillaume Lample,
  L{\'e}lio~Renard Lavaud, Lucile Saulnier, Marie-Anne Lachaux, Pierre Stock,
  Sandeep Subramanian, Sophia Yang, Szymon Antoniak, Teven~Le Scao,
  Th{\'e}ophile Gervet, Thibaut Lavril, Thomas Wang, Timoth{\'e}e Lacroix, and
  William~El Sayed.
\newblock Mixtral of experts.
\newblock \emph{ArXiv}, abs/2401.04088, 2024.
\newblock URL \url{https://api.semanticscholar.org/CorpusID:266844877}.

\bibitem[Johnson(1944)]{Johnson1944Studies}
as~editor Johnson, Wendell.
\newblock Studies in language behavior: A program of research.
\newblock \emph{Psychological Monographs}, 56\penalty0 (2):\penalty0 1--15,
  1944.

\bibitem[Joshi et~al.(2017)Joshi, Choi, Weld, and Zettlemoyer]{TQA2017}
Mandar Joshi, Eunsol Choi, Daniel Weld, and Luke Zettlemoyer.
\newblock Triviaqa: A large scale distantly supervised challenge dataset for
  reading comprehension.
\newblock 05 2017.
\newblock \doi{10.48550/arXiv.1705.03551}.

\bibitem[Jégou et~al.(2011)Jégou, Douze, and Schmid]{JeDouSch2011}
Herve Jégou, Matthijs Douze, and Cordelia Schmid.
\newblock Product quantization for nearest neighbor search.
\newblock \emph{IEEE Transactions on Pattern Analysis and Machine
  Intelligence}, 33\penalty0 (1):\penalty0 117--128, 2011.
\newblock \doi{10.1109/TPAMI.2010.57}.

\bibitem[Klambauer et~al.(2017)Klambauer, Unterthiner, Mayr, and
  Hochreiter]{Klambauer2017SelfNormalizingNN}
G{\"u}nter Klambauer, Thomas Unterthiner, Andreas Mayr, and Sepp Hochreiter.
\newblock Self-normalizing neural networks.
\newblock In \emph{Neural Information Processing Systems}, 2017.
\newblock URL \url{https://api.semanticscholar.org/CorpusID:13713980}.

\bibitem[Kwiatkowski et~al.(2019)Kwiatkowski, Palomaki, Redfield, Collins,
  Parikh, Alberti, Epstein, Polosukhin, Devlin, Lee, Toutanova, Jones, Kelcey,
  Chang, Dai, Uszkoreit, Le, and Petrov]{kwiatkowski-etal-2019-natural}
Tom Kwiatkowski, Jennimaria Palomaki, Olivia Redfield, Michael Collins, Ankur
  Parikh, Chris Alberti, Danielle Epstein, Illia Polosukhin, Jacob Devlin,
  Kenton Lee, Kristina Toutanova, Llion Jones, Matthew Kelcey, Ming-Wei Chang,
  Andrew~M. Dai, Jakob Uszkoreit, Quoc Le, and Slav Petrov.
\newblock Natural questions: A benchmark for question answering research.
\newblock \emph{Transactions of the Association for Computational Linguistics},
  7:\penalty0 452--466, 2019.
\newblock \doi{10.1162/tacl_a_00276}.
\newblock URL \url{https://aclanthology.org/Q19-1026}.

\bibitem[Lai et~al.(2017)Lai, Xie, Liu, Yang, and Hovy]{Lai2017RACELR}
Guokun Lai, Qizhe Xie, Hanxiao Liu, Yiming Yang, and Eduard~H. Hovy.
\newblock Race: Large-scale reading comprehension dataset from examinations.
\newblock In \emph{Conference on Empirical Methods in Natural Language
  Processing}, 2017.
\newblock URL \url{https://api.semanticscholar.org/CorpusID:6826032}.

\bibitem[Lample et~al.(2019)Lample, Sablayrolles, Ranzato, Denoyer, and
  J{\'e}gou]{lample2019large}
Guillaume Lample, Alexandre Sablayrolles, Marc'Aurelio Ranzato, Ludovic
  Denoyer, and Herv{\'e} J{\'e}gou.
\newblock Large memory layers with product keys.
\newblock \emph{Advances in Neural Information Processing Systems}, 32, 2019.

\bibitem[Lee et~al.(2022)Lee, Yang, Lee, and Hwang]{Leeetal2022}
Kyungsu Lee, Jaeseung Yang, Haeyun Lee, and Jae~Youn Hwang.
\newblock Stochastic adaptive activation function.
\newblock In \emph{Proceedings of the 36th International Conference on Neural
  Information Processing Systems}, NIPS '22, Red Hook, NY, USA, 2022. Curran
  Associates Inc.
\newblock ISBN 9781713871088.

\bibitem[Liu et~al.(2024)Liu, Ponnusamy, Cai, Guo, Kim, and
  Athiwaratkun]{liu2024training}
James Liu, Pragaash Ponnusamy, Tianle Cai, Han Guo, Yoon Kim, and Ben
  Athiwaratkun.
\newblock Training-free activation sparsity in large language models.
\newblock \emph{arXiv preprint arXiv:2408.14690}, 2024.

\bibitem[Liu et~al.(2023{\natexlab{a}})Liu, Xia, Wang, and Zhang]{evalplus}
Jiawei Liu, Chunqiu~Steven Xia, Yuyao Wang, and Lingming Zhang.
\newblock Is your code generated by chat{GPT} really correct? rigorous
  evaluation of large language models for code generation.
\newblock In \emph{Thirty-seventh Conference on Neural Information Processing
  Systems}, 2023{\natexlab{a}}.
\newblock URL \url{https://openreview.net/forum?id=1qvx610Cu7}.

\bibitem[Liu et~al.(2023{\natexlab{b}})Liu, Wang, Dao, Zhou, Yuan, Song,
  Shrivastava, Zhang, Tian, Re, et~al.]{liu2023deja}
Zichang Liu, Jue Wang, Tri Dao, Tianyi Zhou, Binhang Yuan, Zhao Song, Anshumali
  Shrivastava, Ce~Zhang, Yuandong Tian, Christopher Re, et~al.
\newblock Deja vu: Contextual sparsity for efficient llms at inference time.
\newblock In \emph{International Conference on Machine Learning}, pp.\
  22137--22176. PMLR, 2023{\natexlab{b}}.

\bibitem[Loshchilov \& Hutter(2017)Loshchilov and
  Hutter]{Loshchilov2017DecoupledWD}
Ilya Loshchilov and Frank Hutter.
\newblock Decoupled weight decay regularization.
\newblock In \emph{International Conference on Learning Representations}, 2017.
\newblock URL \url{https://api.semanticscholar.org/CorpusID:53592270}.

\bibitem[Maas et~al.(2013)Maas, Hannun, Ng, et~al.]{maas2013rectifier}
Andrew~L Maas, Awni~Y Hannun, Andrew~Y Ng, et~al.
\newblock Rectifier nonlinearities improve neural network acoustic models.
\newblock In \emph{Proc. icml}, volume~30, pp.\ ~3. Atlanta, GA, 2013.

\bibitem[Mihaylov et~al.(2018)Mihaylov, Clark, Khot, and
  Sabharwal]{mihaylov-etal-2018-suit}
Todor Mihaylov, Peter Clark, Tushar Khot, and Ashish Sabharwal.
\newblock Can a suit of armor conduct electricity? a new dataset for open book
  question answering.
\newblock In Ellen Riloff, David Chiang, Julia Hockenmaier, and Jun{'}ichi
  Tsujii (eds.), \emph{Proceedings of the 2018 Conference on Empirical Methods
  in Natural Language Processing}, pp.\  2381--2391, Brussels, Belgium,
  October-November 2018. Association for Computational Linguistics.
\newblock \doi{10.18653/v1/D18-1260}.
\newblock URL \url{https://aclanthology.org/D18-1260/}.

\bibitem[Mu \& Lin(2025)Mu and Lin]{MuLin2025}
Siyuan Mu and Sen Lin.
\newblock A comprehensive survey of mixture-of-experts: Algorithms, theory, and
  applications, 03 2025.

\bibitem[Ramachandran et~al.(2018)Ramachandran, Zoph, and
  Le]{Ramachandran2018SearchingFA}
Prajit Ramachandran, Barret Zoph, and Quoc~V. Le.
\newblock Searching for activation functions.
\newblock \emph{ArXiv}, abs/1710.05941, 2018.
\newblock URL \url{https://api.semanticscholar.org/CorpusID:10919244}.

\bibitem[Redmon et~al.(2015)Redmon, Divvala, Girshick, and
  Farhadi]{Redmon2015YouOL}
Joseph Redmon, Santosh~Kumar Divvala, Ross~B. Girshick, and Ali Farhadi.
\newblock You only look once: Unified, real-time object detection.
\newblock \emph{2016 IEEE Conference on Computer Vision and Pattern Recognition
  (CVPR)}, pp.\  779--788, 2015.
\newblock URL \url{https://api.semanticscholar.org/CorpusID:206594738}.

\bibitem[Ridnik et~al.(2021)Ridnik, Lawen, Noy, Ben, Sharir, and
  Friedman]{Ridniketal2021}
Tal Ridnik, Hussam Lawen, Asaf Noy, Emanuel Ben, Baruch~Gilad Sharir, and
  Itamar Friedman.
\newblock Tresnet: High performance gpu-dedicated architecture.
\newblock In \emph{2021 IEEE Winter Conference on Applications of Computer
  Vision (WACV)}, pp.\  1399--1408, 2021.
\newblock \doi{10.1109/WACV48630.2021.00144}.

\bibitem[Sakaguchi et~al.(2020)Sakaguchi, Le~Bras, Bhagavatula, and
  Choi]{Sakaguchi_LeBras_Bhagavatula_Choi_2020}
Keisuke Sakaguchi, Ronan Le~Bras, Chandra Bhagavatula, and Yejin Choi.
\newblock Winogrande: An adversarial winograd schema challenge at scale.
\newblock \emph{Proceedings of the AAAI Conference on Artificial Intelligence},
  34\penalty0 (05):\penalty0 8732--8740, Apr. 2020.
\newblock \doi{10.1609/aaai.v34i05.6399}.
\newblock URL \url{https://ojs.aaai.org/index.php/AAAI/article/view/6399}.

\bibitem[Shazeer(2020)]{Shazeer2020GLUVI}
Noam~M. Shazeer.
\newblock Glu variants improve transformer.
\newblock \emph{ArXiv}, abs/2002.05202, 2020.
\newblock URL \url{https://api.semanticscholar.org/CorpusID:211096588}.

\bibitem[Srivastava et~al.(2014)Srivastava, Hinton, Krizhevsky, Sutskever, and
  Salakhutdinov]{srivastava2014dropout}
Nitish Srivastava, Geoffrey Hinton, Alex Krizhevsky, Ilya Sutskever, and Ruslan
  Salakhutdinov.
\newblock Dropout: a simple way to prevent neural networks from overfitting.
\newblock \emph{The journal of machine learning research}, 15\penalty0
  (1):\penalty0 1929--1958, 2014.

\bibitem[Su et~al.(2021)Su, Lu, Pan, Wen, and Liu]{Su2021RoFormerET}
Jianlin Su, Yu~Lu, Shengfeng Pan, Bo~Wen, and Yunfeng Liu.
\newblock Roformer: Enhanced transformer with rotary position embedding.
\newblock \emph{ArXiv}, abs/2104.09864, 2021.
\newblock URL \url{https://api.semanticscholar.org/CorpusID:233307138}.

\bibitem[Touvron et~al.(2023)Touvron, Lavril, Izacard, Martinet, Lachaux,
  Lacroix, Rozi{\`e}re, Goyal, Hambro, Azhar, Rodriguez, Joulin, Grave, and
  Lample]{Touvron2023LLaMAOA}
Hugo Touvron, Thibaut Lavril, Gautier Izacard, Xavier Martinet, Marie-Anne
  Lachaux, Timoth{\'e}e Lacroix, Baptiste Rozi{\`e}re, Naman Goyal, Eric
  Hambro, Faisal Azhar, Aur'elien Rodriguez, Armand Joulin, Edouard Grave, and
  Guillaume Lample.
\newblock Llama: Open and efficient foundation language models.
\newblock \emph{ArXiv}, abs/2302.13971, 2023.
\newblock URL \url{https://api.semanticscholar.org/CorpusID:257219404}.

\bibitem[Vaswani et~al.(2017)Vaswani, Shazeer, Parmar, Uszkoreit, Jones, Gomez,
  Kaiser, and Polosukhin]{Vaswani2017AttentionIA}
Ashish Vaswani, Noam~M. Shazeer, Niki Parmar, Jakob Uszkoreit, Llion Jones,
  Aidan~N. Gomez, Lukasz Kaiser, and Illia Polosukhin.
\newblock Attention is all you need.
\newblock In \emph{Neural Information Processing Systems}, 2017.
\newblock URL \url{https://api.semanticscholar.org/CorpusID:13756489}.

\bibitem[Wei et~al.(2024)Wei, Zhu, Zhao, Cheng, Li, Lü, Cheng, Zhang, Zhang,
  Zeng, Wang, Ma, Hu, Yan, Fang, and Zhou]{wei2024skyworkmoedeepdivetraining}
Tianwen Wei, Bo~Zhu, Liang Zhao, Cheng Cheng, Biye Li, Weiwei Lü, Peng Cheng,
  Jianhao Zhang, Xiaoyu Zhang, Liang Zeng, Xiaokun Wang, Yutuan Ma, Rui Hu,
  Shuicheng Yan, Han Fang, and Yahui Zhou.
\newblock Skywork-moe: A deep dive into training techniques for
  mixture-of-experts language models, 2024.
\newblock URL \url{https://arxiv.org/abs/2406.06563}.

\bibitem[Yang et~al.(2024)Yang, Yang, Zhang, Hui, Zheng, Yu, Li, Liu, Huang,
  Dong, Wei, Lin, Yang, Tu, Zhang, Yang, Yang, Zhou, Lin, Dang, Lu, Bao, Yang,
  Yu, Li, Xue, Zhang, Zhu, Men, Lin, Li, Xia, Ren, Ren, Fan, Su, Zhang, Wan,
  Liu, Cui, Zhang, Qiu, Quan, and Wang]{Yang2024Qwen25TR}
Qwen~An Yang, Baosong Yang, Beichen Zhang, Binyuan Hui, Bo~Zheng, Bowen Yu,
  Chengyuan Li, Dayiheng Liu, Fei Huang, Guanting Dong, Haoran Wei, Huan Lin,
  Jian Yang, Jianhong Tu, Jianwei Zhang, Jianxin Yang, Jiaxin Yang, Jingren
  Zhou, Junyang Lin, Kai Dang, Keming Lu, Keqin Bao, Kexin Yang, Le~Yu, Mei Li,
  Mingfeng Xue, Pei Zhang, Qin Zhu, Rui Men, Runji Lin, Tianhao Li, Tingyu Xia,
  Xingzhang Ren, Xuancheng Ren, Yang Fan, Yang Su, Yi-Chao Zhang, Yunyang Wan,
  Yuqi Liu, Zeyu Cui, Zhenru Zhang, Zihan Qiu, Shanghaoran Quan, and Zekun
  Wang.
\newblock Qwen2.5 technical report.
\newblock \emph{ArXiv}, abs/2412.15115, 2024.
\newblock URL \url{https://api.semanticscholar.org/CorpusID:274859421}.

\bibitem[Zellers et~al.(2019)Zellers, Holtzman, Bisk, Farhadi, and
  Choi]{Zellers2019HellaSwagCA}
Rowan Zellers, Ari Holtzman, Yonatan Bisk, Ali Farhadi, and Yejin Choi.
\newblock Hellaswag: Can a machine really finish your sentence?
\newblock In \emph{Annual Meeting of the Association for Computational
  Linguistics}, 2019.
\newblock URL \url{https://api.semanticscholar.org/CorpusID:159041722}.

\bibitem[Zhang \& Sennrich(2019)Zhang and Sennrich]{Zhang2019RootMS}
Biao Zhang and Rico Sennrich.
\newblock Root mean square layer normalization.
\newblock \emph{ArXiv}, abs/1910.07467, 2019.
\newblock URL \url{https://api.semanticscholar.org/CorpusID:113405151}.

\end{thebibliography}
\bibliographystyle{iclr2025_conference}

\appendix

\begin{center}
\LARGE \textsc{Appendix}
\end{center}

In this appendix, we add some extra details to complement the main paper. 
Appendix~\ref{sec:architecture} gives architectural details for the models used in the paper. 
Appendix~\ref{sec:sparsity} gives additional details of the sparsity produced by the models that use the \stocha and \swift activations. Appendix~\ref{sec:multigen_setup} provides the experimental details for the diversity of generations ablation. Appendix~\ref{sec:multiforward} provides the experimental details for the multiple forward passes per generated token ablation.
Appendix~\ref{sec:cpt} gives some details for the CPT experiment with \stocha and \swift.
Appendix~\ref{sec:tanhrelu} discusses how to generalize the \stocha and \swift methods for any pair of non-sparse and sparse activations.
Appendix~\ref{sec:benchmarks} gives the references for all the benchmarks used in the experiments. 
The final Appendix~\ref{sec:multiple_generation} gives some examples of multiple answers generated by \stocha.

\section{Architecture detail}
\label{sec:architecture}

The model architectures we use in the paper are loosely inspired by Llama 3.
Table~\ref{tab:architecture} summarizes their main parameters.

\begin{table}[ht]
\centering
{\small
\begin{tabular}{lll}
\toprule
\textbf{Parameter} & \textbf{\amaiasmall}& \textbf{\amaiamed} \\
\midrule
Number of parameters &1.5B& 3B\\
Layers &28 &36\\
Hidden dimension &1536&2048 \\
Intermediate dimension &8960&11008 \\
Number of attention heads &12& 16\\
Number of key-value heads & 2&2\\
\bottomrule
\end{tabular}}
\caption{\amaiasmall and \amaiamed model parameters}\label{tab:architecture}
\end{table}

\section{Sparsity Analysis}
\label{sec:sparsity}

\begin{table}[b]
\centering
\begin{tabular}{lccccc|c}
\toprule
 & \makebox{train} & \makebox{inference} & \swift & \stocha: p & \swift: $\alpha$ & sparsity (\%) \\
\midrule
\arrayrulecolor{gray!50} 
baselines       & \silu          & \silu &  \xmark          &     - &   &  \phantom{0000}0.0002\\
                & \relu          & \relu &         \xmark   &  -    &   &  94.8 \\
\midrule      
\swift          & \silu          & \relu & \cmark &  -    & 0.05  &  79.9 \\
\midrule
\stocha         & \stodoublesilu & \relu & \checkmark & 0.3 & 0.05 & 88.5 \\
+\swift         & \stodoublesilu & \relu & \checkmark & 0.5 & 0.05 & 86.5 \\
        & \stodoublesilu & \relu & \checkmark & 0.7 & 0.05 & 84.6 \\
\arrayrulecolor{black} 
\bottomrule 
\end{tabular}
\caption{We report the rate of the 0-valued activation after the \silu (not sparse) and \relu activation, when training with the \amaiasmall model with the following activations:
\relu, \silu, \swift fine-tuning, and when using \stocha with \swift for varying values of $p$.  
\label{tab:sparsity}}
\end{table}

\begin{figure}[h!]
\vspace{-1em}
    \centering
    \begin{tabular}{cc}
    Pre-activation inputs distribution & Presence of dead neurons in the $W_1$ layer  \\
    \includegraphics[width=0.45\linewidth]{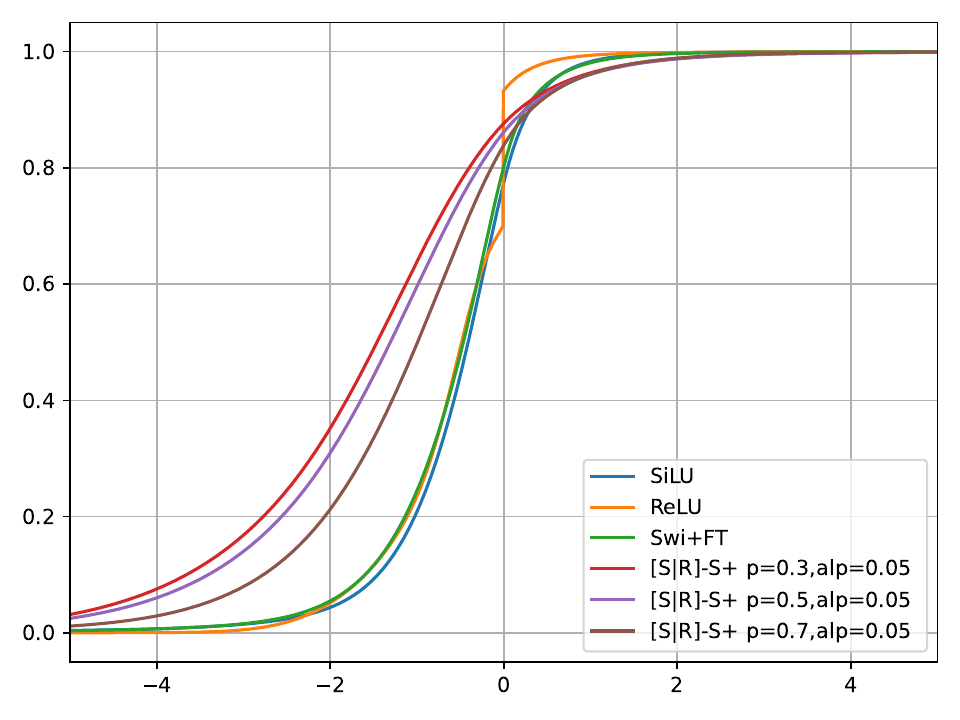} & 
    \includegraphics[width=0.45\linewidth]{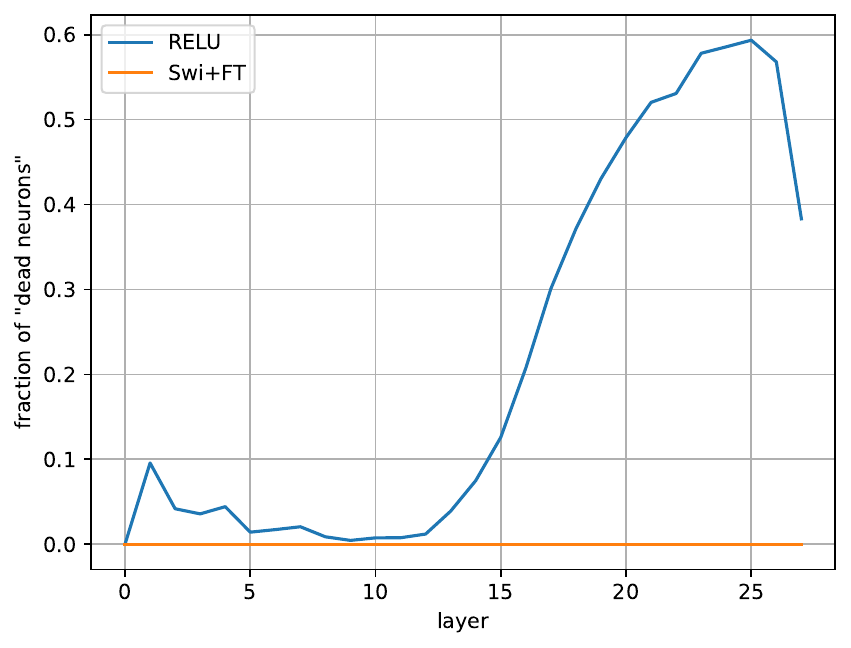} 
    \end{tabular}
\vspace{-1em}    
    \caption{
    \emph{Left:} empirical cumulative distribution functions for the inputs to the activation functions when training the \amaiasmall model with one of the following activations: \silu, \relu, \swift fine-tuning and the combination of \stocha with \swift for different values of $p$.
    \emph{Right:} fraction of near-zero rows of the $W_1$ matrix, indicating useless neurons for two types of activations.
    }
    \label{fig:cdfs}
\end{figure}

In Table~\ref{tab:sparsity}, we report the sparsity ratios resulting from the gating in the FFN layer, when using the following activations:
\relu, \silu, \swift fine-tuning, and when using \stocha with \swift for varying values $p$. In all cases, except for the \silu baseline, we use \relu at test time. We observe that \swift fine-tuning with \relu brings back a lot of sparsity (80\%), yet the model is not competitive. The combination of \stocha with \swift leads to models offering a high sparsity degree by using \relu at test time and simultaneously achieving competitive performance, see Tables~\ref{tab:nll} and \ref{tab:benchmark_detail3B}. 

In Figure~\ref{fig:cdfs} (\emph{left}), we observe that $P(X\leq 0)$ is the highest for \relu since most of its mass is concentrated around zero as in Table~\ref{tab:sparsity}. For $x<0$, the probability $P(X\leq x)$  is lower for \silu than for both \swift and when we combine \stocha with \swift for different values of $p$. This shows that combining \stocha with \swift enables us to control how much mass we assign to negative values of $x$ depending on $p$.

\section{Diversity of generations ablation experimental setup}
\label{sec:multigen_setup}
We use the \amaiamed models trained with the \stocha activation \stodoublesilu with $p$ in $\{0.3, 0.5, 0.7, 0.85\}$. At inference time, the probability $p$ is set to be the same as the one during training. We generate twenty answers for the NQ and TQA generation tasks, compute the Ff1 and Type-token ratio (TTR) metrics of $n$ generated answers (samples) and average over all datapoints for a given number of samples $n$ in $\{2,3,4,5,10,20\}$.

In Figure~\ref{fig:mutiplegen} (b) we report the type-token ratio metric~\citep{Johnson1944Studies}, a simple and widely used metric in linguistics and natural language processing to measure lexical diversity in a text. In this context, a \textit{type} consists of a unique word in the text and a \textit{token} consists of any word occurrence in the text (including repetitions). The Type-Token Ratio is then \textit{TTR=Number of Types / Number of Tokens}.

\section{Multiple forward passes per generated token}
\label{sec:multiforward}

If we use the \stocha activation function, it is equivalent to doing Monte Carlo dropout~\citep{GalGhah2016} on the negative side of the input after activation: it is set to zero with probability $1-p$. For this reason, at test time, if we run multiple forward passes, this naturally creates a random output after activation which is either zero or \silu on the negative side of the inputs after activation. 

 In this setup, at generation time, after performing the multiple forward passes, we select the next token according to the following strategies:

\textbf{Max probability:} we select the token that has the highest probability. 

\textbf{Majority vote:} we select the next token based on the token which appears the most often across the multiple passes.

\textbf{Results.} We use ~\amaiamed models trained with the~\stodoublesilu activations with $p$ in $\left\{0.3, 0.5, 0.7, 0.85\right\}$. We vary the number of forward passes in $\left\{5,10,15,20,30\right\}$. We report the average performance of the generation tasks \texttt{nq, tqa, mbpp} and \texttt{human eval plus} and greedy decoding. Our findings indicate that even though we observe more diversity at the token level, the majority vote and maximum probability strategies do not perform better than the baseline, where we have a single forward pass as we increase the number of forward passes.

\section{Continuous Pretraining \amaiasmall models}
\label{sec:cpt}
We can leverage the \stocha and \swift methods for continuous pretraining (CPT) language models, besides using these activations to train LMs from scratch. In this experiment, we start from the \amaiasmall models pretrained with the 
\silu activation. During CPT, we use i) \swift  ii)
\stocha with $p=0.3$ or $p=0.5$ or iii) \stocha and \swift with $p=0.5$ and $\alpha=0.1$. We use the same number of steps used in the pretraining experiments: 45,629 steps. We compare the training losses against the baseline model that keeps \silu during CPT.

 \begin{table}[h]
\centering
\resizebox{\textwidth}{!}{%
\begin{tabular}{lcccc|c|c}
\toprule
  & \makebox{pretrain} & \makebox{CPT} & \makebox{inference} &  \stocha: p &  sparsity (\%) & train loss \\
\midrule
\arrayrulecolor{gray!50} 
baseline & \silu & \silu & \silu & - &  0.0002 &\textbf{2.076}  \\
  \midrule
  \swift & \silu & \relu & \relu &  - &  \textbf{90.96} & \underline{2.086}  \\
  \midrule
 \stocha & \silu & \stocha & \relu &  0.5 &  41.02  & 2.095 \\
  & \silu & \stocha & \relu &  0.3 &   57.93 & 2.095 \\
  \midrule
 \stocha+\swift & \silu & \stocha+\swift & \relu &  0.5 &  \underline{81.61} & 2.096 \\
\bottomrule
\end{tabular}}
\caption{We report the rate of the 0-valued activations  and training losses when doing CPT with the \amaiasmall model with the following activations:
\relu, \silu, \swift,\stocha and when using \stocha with \swift for varying values $p$.}
\label{tab:cpt}
\end{table}

\section{\stocha and \swift for (\tanh, \relu) activations}
\label{sec:tanhrelu}
Our \stocha and \swift methods can be generalized, starting with any pair of non-sparse, sparse activations, in the following way:
\begin{enumerate}
\item \stocha: the non-sparse (activation which takes non-zero values on the negative side) and sparse activation pair can be used as a stochastic activation function that takes the form of either of the activations on the positive side, and, on the negative side, it either takes the non-zero value of the non-sparse variant with probability $p$ or is set to zero with probability $1-p$.

\item \swift: we start with an LM with a non-sparse activation and train it for $(1-\alpha)\%$ steps. Successively, we finetune the LM with the sparse activation for $\alpha\%$ of the final steps.
\end{enumerate}
To illustrate how the \stocha and \swift methods are generalizable, we use the pair of (\tanh, \relu) activations. We pretrain a \amaiasmall with \tanh as the non-sparse activation and \relu as the sparse one for 1) and 2). We compare these versus the losses to the corresponding baselines: using only the \tanh or only the \relu activations, respectively. We denote the activation for 1) as \tanhft with parameter $\alpha$ and for 2), with \stodoubletanh with parameter $p$.

We observe that there is barely any gap between the train loss of the non-sparse (\tanh, bolded) and the sparse (\relu, in red) activations, hence, there is no trade-off between sparsity and performance, in contrast to the (\silu, \relu) case. 

\begin{table}
\centering
\resizebox{0.8\textwidth}{!}{
\begin{tabular}{l|ccccc|ccc}
\toprule
& \multicolumn{4}{c}{Training activation} & & \multicolumn{3}{c}{\amaiasmall} \\
& \multicolumn{2}{c}{$x<0$} & $x>0$ & $p$ & & train & val & val \\
\cmidrule(lr){2-3}\cmidrule(lr){4-4} 
Activation & $p$ & $1-p$ & & & \swift & & \relu & \stocha \\
\midrule
\tanh & \multicolumn{2}{c}{\cTANH} & \cTANH & - & \xmark & 2.133 & 2.155 & \\
\relu & \multicolumn{2}{c}{\cReLU} & \cReLU & - & \xmark & 2.140 & 2.162 & \\
\tanhft & \multicolumn{2}{c}{\cTANH} & \cTANH & - & \cmark & 2.193 & 2.224 & \\ 
\midrule  
\stodoubletanh & \cTANH & \cReLU & \cTANH & 0.3 & \xmark & 2.368 & 10.803 & 2.421 \\
\stodoubletanh & \cTANH & \cReLU & \cTANH & 0.5 & \xmark & 2.314 & 10.997 & 2.359 \\
\bottomrule
\end{tabular}
}
\caption{Losses: train is computed over the last 500 steps of the training loss of~\amaiasmall, val is measured after training on a different set of text and code using the \relu activation$^\star$ or \stocha, i.e., the same activation used at train time (possibly deterministic). If \swift is enabled, we switch to \relu for the last $10\%$ steps.}
\label{tab:nll_tanhrelu}
\end{table}

\section{Benchmarks}
\label{sec:benchmarks}
\paragraph{Code generation}
We use two benchmarks that evaluate the code generation capabilities of AI models: HumanEval+ and MBPP. 
\begin{itemize} 
    \item The HumanEval+~\citep{evalplus} benchmark is an extension of HumanEval~\citep{chen2021codex}, which is designed to evaluate the functional correctness of code generated by AI models.
    \item MBPP~\citep{Austin2021ProgramSW} is designed to evaluate the code generation abilities of AI models, particularly for Python programming tasks.
\end{itemize} 
\begin{table}[h!]
\centering
\begin{tabular}{|llll|}
\hline
\textbf{Benchmark} & \textbf{Metric} & \textbf{Few Shot} & \textbf{Type} \\
\hline
hellaswag        & acc\_char      &     0              &    choice          \\
winogrande       & acc\_char      &      0             &       choice       \\
arc\_easy        & acc\_char      &       0            &       choice        \\
arc\_challenge   & acc\_char      &      0             &      choice        \\
piqa             & acc\_char      &       0            &    choice         \\
obqa             & acc\_char      &      0             &          choice    \\
race.middle      & acc\_char      &        0           &     choice         \\
race.high        & acc\_char      &       0            &     choice         \\
human\_eval\_plus & pass@1     &     0              &     generation         \\
mbpp             & compiles@1         &      3             &        generation      \\
tqa              & f1             &       5            &      generation        \\
nq               & f1             &         5          &   generation           \\
\hline
\end{tabular}
    \caption{Summary of benchmarks used for evaluation. We few-shot prompt the~\amaiasmall and~\amaiamed models. The metric depends on the benchmark as well as the average performance. \label{tab:benchmarksconfigs}}
\end{table}

\paragraph{Common sense and general reasoning}  We use benchmarks consisting of question-answer or multiple-choice questions designed to evaluate the commonsense reasoning abilities of AI models, particularly in the context of natural language understanding: HellaSWAG, ARC, PIQA, OBQA, Winogrande, NaturalQuestions, RACE, TQA and GSM8K. 

\begin{itemize}
    \item HellaSWAG~\citep{Zellers2019HellaSwagCA} consists of multiple-choice questions where each question contains a short context (a sentence or paragraph) followed by four possible continuations. Only one continuation is correct and makes sense given the context.
    \item The AI Reasoning Challenge (ARC)~\citep{Clark2018ThinkYH} benchmark consists of multiple-choice science questions typically found in elementary and middle school exams. The ARC questions require a mix of factual knowledge, commonsense reasoning, and multi-step inference.
    \item The Physical Interaction Question-Answering (PIQA)~\citep{Bisk_Zellers_Lebras_Gao_Choi_2020} benchmark consists of multiple-choice questions about how to accomplish simple physical tasks. Each question presents a short scenario and two possible solutions; only one is physically plausible.
    \item The OpenBook QA (OBQA)~\citep{mihaylov-etal-2018-suit} benchmark consists of 6,000 multiple-choice questions based on elementary science facts. Each question is designed to require combining a provided ``open book'' science fact with additional commonsense or general knowledge. 
    \item WinoGrande~\citep{Sakaguchi_LeBras_Bhagavatula_Choi_2020} benchmark consists of multiple-choice questions.
Each question presents a sentence with a pronoun and two possible antecedents; the task is to choose the correct referent for the pronoun.
    \item Natural Questions (NQ)~\citep{kwiatkowski-etal-2019-natural} benchmark is designed to evaluate the ability of an AI model to answer real user questions using information from Wikipedia.
    \item The Reading Comprehension from Examinations (RACE)~\citep{Lai2017RACELR} benchmark consists of passages and multiple-choice questions to assess how well AI models can comprehend and reason about written passages.
    \item The Trivia QA benchmark (TQA)~\citep{TQA2017} is a reading comprehension dataset that pairs trivia questions with evidence documents from which answers can be derived.
    \item The Grade School Math 8k (GSM8k)~\citep{Cobbe2021TrainingVT} benchmark is a dataset of 8,500 high-quality, linguistically diverse grade school math word problems. The benchmark is designed to test multi-step mathematical reasoning capabilities in language models.
    
\end{itemize}

\section{Examples of various sequences generated with \stocha}
\label{sec:multiple_generation}
We use a pre-trained \amaiamed model with \stocha, we can use the \stocha activation at test time to generate multiple predictions by leveraging the randomness from the activation function. In Table~\ref{tab:exmultigen}, we provide two examples of the generations obtained using questions from the TQA benchmark~\citep{TQA2017}:
\newcommand{\EXQA}[2]{\multirow{6}{5cm}{{\bf #1}\\{#2}}}

\begin{longtable}{p{5cm}|rrrr}

\toprule
question and     & generated & number of  & NLL  & F1 \\
accepted answers & answers & times & score & score \\
\midrule
\endfirsthead

\multicolumn{5}{c}{\tablename\ \thetable\ -- Continued from previous page} \\
\toprule
question and     & generated & number of  & NLL  & F1 \\
accepted answers & answers & times & score & score \\
\midrule
\endhead

\midrule
\multicolumn{5}{r}{Continued on next page} \\
\endfoot

\bottomrule
\endlastfoot

\EXQA{When did ibuprofen become available over the counter?}{A: 1983\\ A: 1984}
 &  1988 & 4 & 1.002 & 0 \\*
&    1982 & 11 & 1.004 & 0 \\*
&    1980 & 1 & 1.030 & 0 \\*
&    1995 & 2 & 1.019 & 0 \\*
&    2001 & 1 & 1.017 & 0 \\*
&    1983 & 1 & 1.024 & 1 \\
\EXQA{Who played michael jackson in jackson 5 movie?}{A:alex burrall\\A:abolade david olatunde\\A:wylie draper\\A:jason weaver\\}
   &  Michael Jackson & 19 & 0.948 & 0 \\*
   &  Michael jackson & 1 & 0.792 & 0 \\*
   &&& \\*
   &&& \\*
   &&& \\*
   &&& \\
\midrule
\EXQA{Where is the meridian that is opposite the prime meridian located?}{A:180th meridian\\A:antimeridian\\}
   &  Greenwich & 9 & 0.896 & 0 \\*
   &  Greenwich, England & 11 & 0.551 & 0 \\*
   &&& \\*
   &&& \\*
   &&& \\
\midrule
\EXQA{The cold dry winds that blow over northern india in winter are called?}{A:northeast monsoon\\A:retreating monsoon\\A:northeast monsoon or retreating monsoon\\}
   &  Siberian & 7 & 1.099 & 0 \\*
   &  Siberian winds & 6 & 0.970 & 0 \\*
   &  Ganges & 1 & 1.617 & 0 \\*
   &  monsoon & 2 & 1.051 & 2/3 \\*
   &  katabatic winds & 3 & 0.738 & 0 \\*
   &  Ganga & 1 & 1.643 & 0 \\*
   &&& \\
\midrule
\EXQA{Where can you find dna in the body?}{A:chromosomes in cell\\A:inside cell nucleus\\}
   &  Blood & 6 & 1.262 & 0 \\*
   &  Hair & 12 & 1.393 & 0 \\*
   &  Mitochondria & 1 & 0.700 & 0 \\*
   &  In the nucleus & 1 & 1.437 & 2/5 \\*
&&& \\
\midrule
\EXQA{Who is often associated with printing the first book using moveable type in germany?}{A:johannes gutenberg\\}
   &  Johannes Gutenberg & 20 & 0.205 & 1 \\*
      &&& \\*
   &&& \\*
   &&& \\*
   &&& \\
\midrule
\EXQA{Who won the womens 2017 ncaa basketball tournament?}{A:south carolina\\}
   &  Kentucky & 15 & 1.852 & 0 \\*
   &  Kentucky Wildcats & 1 & 1.304 & 0 \\*
   &  North Carolina & 4 & 1.342 & 1/2 \\*
   &&& \\*
   &&& \\   
\midrule
\EXQA{Country with most olympic gold medals all time?}{A:united states\\}
   &  USA & 14 & 0.818 & 0 \\*
   &  United States & 6 & 0.614 & 1 \\*
   &&& \\*
   &&& \\   
\midrule
\EXQA{The atomic number of indium which belongs to 5th period is?}{A:49\\}
   &  5 & 14 & 0.579 & 0 \\*
   &  49 & 5 & 0.650 & 1 \\*
   &  84 & 1 & 0.695 & 0 \\*
   &&& \\*
   &&& \\   
\midrule
\EXQA{Who appoints the members of the board of governors of the federal reserve?}{A:president\\}
   &  The president & 16 & 0.602 & 1 \\*
   &  The president of the United States & 4 & 0.431 & 2/5 \\*
  &&& \\*
   &&& \\ 
\EXQA{What age do you need to be to buy a bb gun?}{A:18\\}
   &  14 & 17 & 0.660 & 0 \\*
   &  10 & 2 & 0.680 & 0 \\*
   &  18 & 1 & 0.728 & 1 \\*
   &&& \\   
\midrule
\EXQA{What genre is the magic tree house books?}{A:childrens historical fantasy}
   &  Fantasy & 14 & 1.038 & 1/2 \\*
   &  Children's fiction & 1 & 1.051 & 2/5 \\*
   &  Children's & 3 & 1.245 & 1/2 \\*
   &  Children's books & 2 & 1.067 & 2/5 \\
   &&& \\   
\midrule
\EXQA{What is the name of the skin between your nostrils?}{A:nasal septum\\A:septum\\}
   &  Nasal septum & 17 & 0.707 & 1 \\*
   &  Nasion & 3 & 1.224 & 0 \\*
   &&& \\*
   &&& \\ *
    &&& \\ 
   \midrule
    \caption{Example generations. For each question, we indicate the ground-truth answers (from the dataset). 
    We generate 20 answers per question with \texttt{[S|R]-S+}, $p$=0.7. 
    We list the de-duplicated answers, with the NLL score (used to sort the results) and the F1 score (used to evaluate the result, it is computed as the intersection of bags-of-words).} 
\label{tab:exmultigen} \\[-3\baselineskip]
\end{longtable}

\end{document}